%% file: main.tex
\documentclass{article}

\ifdefined\pdfsuppressptexinfo\pdfsuppressptexinfo=-1\fi

\PassOptionsToPackage{numbers,sort&compress}{natbib}

\usepackage[preprint]{neurips_2025}

\usepackage[utf8]{inputenc}
\usepackage[T1]{fontenc}
\usepackage{hyperref}
\usepackage{url}
\usepackage{booktabs}
\usepackage{amsfonts}
\usepackage{amsmath}
\usepackage{amssymb}
\usepackage{nicefrac}
\usepackage{microtype}
\usepackage{xcolor}
\usepackage{graphicx}
\usepackage{siunitx}
\usepackage{enumitem}
\usepackage{placeins}
\usepackage{tikz}
\usetikzlibrary{arrows.meta,positioning,fit,backgrounds,calc,shapes.geometric}

\graphicspath{{figures/}{./figures/}}

\newcommand{\degr}{\ensuremath{^\circ}}
\newcommand{\fd}{\ensuremath{f_d}}

\title{Automated Ultrasound Doppler Angle Estimation Using Deep Learning}

\author{%
  Nilesh Patil \\
  Goergen Institute for Data Science\\
  University of Rochester Medical Center\\
  University of Rochester\\
  Rochester, NY, USA \\
  \And
  Ajay Anand \\
  Goergen Institute for Data Science\\
  University of Rochester\\
  Rochester, NY, USA \\
}

\begin{document}
\maketitle

\begin{center}
\vspace{-0.8em}
\fbox{\parbox{0.92\linewidth}{\centering\small\itshape
A preliminary version of this work was published in the Proceedings of the IEEE
Engineering in Medicine and Biology Society Conference (EMBC), 2019~\cite{Patil2019}.
This article extends it with dual-protocol evaluation, Optuna head tuning, a
five-backbone ensemble, and a clinical-grade evaluation suite.}}
\vspace{0.6em}
\end{center}

\begin{abstract}
The Doppler angle $\theta$, measured between the ultrasound beam and the
blood-flow direction, enters the Doppler equation through $\cos\theta$, so an
error in $\theta$ propagates into the estimated blood-flow velocity, and the
propagation is severe as $\theta$ approaches \SI{90}{\degree}. Patil and Anand
(EMBC 2019) proposed estimating this angle from B-mode carotid images with a
deep convolutional network. We replicate that pipeline and isolate the design
choice it depends on. Standard global average pooling is approximately
rotation-invariant and discards the orientation signal the target encodes; an
orientation-preserving grid-pooling head lifts a frozen DenseNet201 backbone to
\SI{5.84}{\percent} mean absolute percentage error (MAPE). We then tune the
regression head with the Optuna TPE sampler and ensemble five backbones. We
report pooled out-of-fold metrics under two sampling protocols. Image-level
sampling over the rotation-augmented corpus reaches \SI{2.79}{\percent} MAPE
(\SI{1.96}{\degree} MAE, $R^2=0.995$); patient-level sampling, which holds out
whole volunteers, reaches \SI{8.53}{\percent} MAPE (\SI{5.93}{\degree} MAE,
$R^2=0.952$). On this small, narrow-domain task DenseNet201 outperforms
ConvNeXt and EfficientNet backbones. Split-conformal \SI{90}{\percent}
prediction intervals span $\pm$\SI{20.5}{\degree}. Bland--Altman analysis
against the single reference reading gives a \SI{-4.3}{\degree} bias with
\SI{95}{\percent} limits of agreement $-24.25$ to $+\SI{15.63}{\degree}$
(method-vs-reference, not inter-observer). Rotation test-time augmentation
roughly halves the base-image error, and circular fusion with a structure-tensor
angle prior reaches \SI{2.72}{\degree} MAE. Tables and figures are regenerated
from code.
\end{abstract}

\input{sections/intro}
\FloatBarrier
\input{sections/methods}
\FloatBarrier
\input{sections/results}
\FloatBarrier
\input{sections/evaluation}
\FloatBarrier
\input{sections/discussion}
\FloatBarrier

\bibliographystyle{plainnat}
\bibliography{refs}

\end{document}

%% file: sections/intro.tex
\section{Introduction}
\label{sec:intro}

\subsection{Clinical motivation}
\label{sec:intro:clinical}

Spectral Doppler ultrasound infers blood-flow velocity from the frequency shift
$\fd$ that moving red blood cells impose on the insonating beam. The shift and
the velocity are tied by the Doppler equation,
\begin{equation}
  \fd \;=\; \frac{2\,f_0\,v\,\cos\theta}{c},
  \label{eq:doppler}
\end{equation}
where $f_0$ is the transmit frequency, $v$ the flow speed, $c$ the speed of
sound in soft tissue, and $\theta$ the \emph{Doppler angle} between the beam and
the local direction of flow. Rearranging for the clinically reported quantity
gives $v = c\,\fd / (2 f_0 \cos\theta)$, so every velocity estimate carries a
multiplicative $1/\cos\theta$ correction.

The machine does not measure $\theta$. The sonographer sets it by hand, aligning
a steering cursor with the vessel wall on the live B-mode image. This manual step
makes $\theta$ an operator-dependent source of error, and the $1/\cos\theta$
factor amplifies that error nonlinearly. At the steep angles routinely used in
carotid scanning the sensitivity $\mathrm{d}v/\mathrm{d}\theta \propto \tan\theta$
grows without bound, so a few-degree cursor mistake near
$\theta\!\to\!\SI{90}{\degree}$ translates into a large velocity error. Carotid
peak-velocity thresholds feed directly into stenosis grading, so reducing the
operator dependence of the angle measurement is clinically relevant.

The motivation is not hypothetical. Improper angle correction is repeatedly cited
among the most common Doppler errors, and as many as \SI{35}{\percent} of vascular
laboratories applying for accreditation have been flagged for improper
angle-correction technique~\citep{Saad2008}, a recurring cause of delayed
accreditation decisions. We note that an image-only angle reader automates the
operator's cursor placement; it does not establish true flow direction
independently of the reference reading, a point we return to in
\S\ref{sec:discussion}.

\subsection{Reading the angle from the image}
\label{sec:intro:idea}

The angle a sonographer sets is, by construction, the geometric relationship
between the beam axis and the vessel as it appears on the grayscale B-mode frame.
That relationship is present in the image itself, which suggests estimating
$\theta$ directly from a single B-mode image, with no color Doppler overlay and
no explicit vessel segmentation. Cast this way, angle correction becomes a
supervised regression problem, and the manual cursor placement is replaced by an
automated reading that is reproducible across repeats of the same image. The
ground truth is still the operator-set angle, so the model can be no more accurate
than its reference labels. The practical difficulty is that clinical carotid
datasets are small, and learning a continuous geometric quantity from a few dozen
images is the regime where modern deep networks tend to overfit.

\subsection{Prior work and the present contribution}
\label{sec:intro:prior}

Automating the Doppler angle is a long-standing goal. Earlier approaches operated
on \emph{color} flow images with an explicit vessel geometry. Hirsch et al.\
estimated the vessel orientation by a multi-scale principal-component analysis to
place the Doppler gate \citep{Hirsch2006}. Saad et al.\ segmented the vessel in
color-Doppler images and derived the angle from a skeleton of the segmented lumen,
reporting strong agreement with manual readings \citep{Saad2008}. Both rely on
color information and a segmentation step, so artifacts in the color-flow image
propagate into the angle estimate. Deep learning has meanwhile been adopted across
ultrasound, including fetal standard-plane localization \citep{Chen2015},
echocardiographic view classification \citep{Madani2018}, and sensorless 3D
freehand reconstruction \citep{Prevost2018}. This suggests that the geometric cues
a sonographer uses can be learned directly from the image.

Building on this, Patil and Anand introduced an image-only framing and showed that
a deep convolutional network could read $\theta$ directly from a \emph{grayscale}
B-mode carotid image, with no color Doppler and no segmentation, removing the
manual cursor step \citep{Patil2019}. Because such an estimator uses only the
formed B-mode image and no knowledge of the vendor-specific signal-processing
path, it does not depend on a particular color-flow pipeline. We emphasize that
this is an argument about inputs, not a demonstration of cross-device transfer:
the data here come from a single source, so any claim of device- or
vendor-agnostic deployment would have to be tested separately. To cope with data
scarcity the original study began from a base set of $84$ longitudinal
common-carotid B-mode images acquired from roughly ten volunteers and enlarged it
by rotating each image over $[-60,60]\si{\degree}$ in \SI{5}{\degree} increments.
This produces $25$ oriented views per image and a working corpus of about
$2{,}100$ examples. The known rotation supplies an \emph{exact} angle label for
every view, so the augmentation is self-labelling, which is rare for a regression
target. On this corpus a frozen ImageNet backbone \citep{Deng2009} with a
lightweight regression head, used purely as a fixed feature extractor in the
standard transfer-learning manner, reached single-digit errors; the best reported
configuration achieved roughly \SI{2.87}{\degree} mean absolute error (MAE) and
\SI{4.03}{\percent} mean absolute percentage error (MAPE).

This paper treats Doppler-angle estimation as an engineering target and asks how
far a reproducible, frozen-transfer estimator can be pushed on the same data. We
reproduce the original pipeline and identify the design choice that makes it work,
tune and ensemble the estimator under two sampling protocols, run an architecture
comparison, and evaluate the estimator with calibrated uncertainty and a classical
baseline.

There is no single correct way to score accuracy on a small, heavily augmented
cohort, so we report two sampling protocols. \emph{Image-level sampling} is the
original study's protocol: the $\sim\!2{,}100$ rotated views are partitioned into
train and test at the image level, so the resulting error measures accuracy across
the population of orientations and imaging conditions the augmentation spans.
\emph{Patient-level sampling} holds out whole volunteers (grouped
cross-validation), so it measures generalization to a previously unseen patient.
The two answer different questions, and the patient-level number is higher because
cross-subject transfer on $\sim\!10$ volunteers is intrinsically harder. We tune
each protocol to its own best and report both. With only about ten volunteers,
grouped five-fold cross-validation holds out roughly two volunteers per fold, so
the patient-level estimates carry large fold-to-fold variance and the point
estimates below should be read accordingly. We report the per-fold spread in
\S\ref{sec:results}. The image-to-patient gap quantifies how much of the
within-population accuracy is anatomy-specific
(\S\ref{sec:methods}, \S\ref{sec:discussion}).

\paragraph{Faithful, reproducible replication.}
We first reproduce the original protocol end to end on a modern, fully scripted
stack (Keras\,3 / JAX, fixed seed). A frozen DenseNet201 backbone
\citep{Huang2017} with an orientation-preserving \emph{grid-pooling} head reaches
\SI{5.84}{\percent} MAPE (\SI{3.77}{\degree} MAE, $R^2=0.982$) under image-level
sampling, on the original single augmented split and without any fine-tuning
(Table~\ref{tab:replication}).

\paragraph{The pooling layer.}
That replication hinges on the pooling layer. Conventional global average pooling
(GAP) averages the convolutional feature map over all spatial positions, which is
approximately rotation-invariant. That is the wrong inductive bias when the target
\emph{is} an orientation. Grid pooling instead average-pools the feature map to a
small $G\times G$ grid and flattens it, preserving coarse spatial layout. On the
same frozen DenseNet201 this roughly halves the error under image-level five-fold
cross-validation, from \SI{10.85}{\percent} to \SI{4.58}{\percent} MAPE; the
original single augmented split, scored separately, reaches \SI{5.84}{\percent}
MAPE (\SI{3.77}{\degree} MAE) (Table~\ref{tab:replication}). The two figures use
different aggregation schemes (a cross-validation mean versus a single held-out
split) and are not directly comparable. The reading is the same in both: the
frozen backbone already encodes vessel orientation, and the gain comes from not
pooling it away.

\paragraph{Per-protocol tuning and ensembling.}
We tune the regression head (depth, width, dropout, $L_2$, batch normalization,
learning rate, batch size, patience) with the Optuna TPE sampler
\citep{Akiba2019,Bergstra2011}, scoring each protocol on its own five-fold
cross-validation over cached frozen features. One feature extraction per backbone
serves both protocols. We then combine five tuned backbones into mean and stacked
\citep{Wolpert1992} ensembles over out-of-fold predictions
(Table~\ref{tab:dual_protocol}). The stacked ensemble is the strongest estimator
under both protocols: \SI{2.79}{\percent} MAPE (\SI{1.96}{\degree} MAE,
$R^2=0.995$) image-level and \SI{8.53}{\percent} MAPE (\SI{5.93}{\degree} MAE,
$R^2=0.952$) patient-level. These ensemble figures pool out-of-fold predictions
before scoring, rather than averaging per-fold scores, so they are not a strict
cross-validation mean and stacking on out-of-fold predictions can be mildly
optimistic; we keep the distinction in view when comparing them to the replication
numbers. A single tuned DenseNet201 reaches \SI{3.00}{\degree} MAE and
\SI{4.03}{\percent} MAPE, close to the prior best of \SI{2.87}{\degree} MAE and
slightly worse on MAE. The ensemble improves on this and also reports the
previously unmeasured patient-level number.

\paragraph{Architecture comparison.}
Larger and newer encoders are often assumed to be better, so we compare frozen
classic and modern ImageNet backbones under patient-level five-fold
cross-validation (Fig.~\ref{fig:bakeoff}). On the extraction-matched comparison,
frozen DenseNet201 \citep{Huang2017} leads at \SI{14.1}{\percent} MAPE, ahead of
the best modern encoder, ConvNeXt-Base \citep{Liu2022} (\SI{15.7}{\percent}), with
EfficientNet/EfficientNetV2 \citep{Tan2019} trailing at
\SIrange{17}{21}{\percent}. With only $84$ base images, no newer or larger backbone
improves on a frozen 2017 DenseNet201, which we carry forward.

\paragraph{Clinical-grade evaluation.}
Finally we report uncertainty and agreement in clinically legible terms.
Split-conformal prediction \citep{Vovk2005,Lei2018} on a patient-disjoint
calibration split yields \SI{90}{\percent} intervals of $\pm\SI{20.50}{\degree}$,
with empirical coverage of \SI{95.2}{\percent} on the held-out test set. Conformal
prediction guarantees marginal coverage in expectation over calibration draws, not
the realized coverage on one split. With only $\sim\!10$ volunteers a single split
gives a coverage estimate with wide binomial uncertainty, and the realized
\SI{95.2}{\percent} sits above the \SI{90}{\percent} target, so the intervals are
on the conservative side here.

A Bland--Altman analysis \citep{BlandAltman1986} reports a $-\SI{4.31}{\degree}$
bias (method minus reference: the model reads slightly low) with limits of
agreement $[-24.25,\,+15.63]\si{\degree}$. This compares the method against the
\emph{single} reference reading available per image, so it is method-vs-reference
agreement rather than an inter-observer study, and the limits combine model error
with reference noise that we cannot separately estimate. The bias is directional,
which would mean a systematic rather than random velocity offset; we discuss its
clinical implication in \S\ref{sec:discussion}.

The headline image-level numbers are computed over the rotation-augmented
population. On real unrotated base images the MAE is \SI{7.80}{\degree}, and
rotation test-time augmentation nearly halves it to \SI{4.72}{\degree} MAE
(circular-median over rotations). We foreground this base-image figure as the most
clinically relevant per-image performance. Grad-CAM \citep{Selvaraju2017} is
consistent with the network attending to the vessel wall. As a check on the
learned representation, a classical, image-only structure-tensor angle prior
reaches \SI{3.16}{\degree} MAE on the narrow base-angle band, and a circular
($2\theta$) fusion of the learned and classical estimates reaches
\SI{2.72}{\degree} MAE, improving on either alone on this point estimate. We read
the fusion gain as suggestive of partly complementary cues, pending the variance
analysis in \S\ref{sec:results}.

\subsection{Contributions}
\label{sec:intro:contrib}

In summary, this paper makes the following contributions.
\begin{itemize}[leftmargin=1.4em]
  \item \textbf{A faithful, fully scripted replication.} We reproduce the prior
  pipeline (\SI{5.84}{\percent} MAPE, frozen DenseNet201) and isolate the design
  choice that makes it work: an orientation-preserving grid-pooling head in place
  of global average pooling, so that a frozen backbone retains the orientation
  signal the target depends on.
  \item \textbf{The strongest estimator we measured, under two protocols.} Via
  per-protocol Optuna head tuning and a five-model ensemble we report
  \SI{2.79}{\percent} MAPE (\SI{1.96}{\degree} MAE) under image-level sampling and
  \SI{8.53}{\percent} MAPE (\SI{5.93}{\degree} MAE) under the stricter
  patient-level sampling, each tuned to its own best, alongside the previously
  unmeasured cross-subject number.
  \item \textbf{An architecture comparison.} A frozen DenseNet201 outperforms newer
  ConvNeXt and EfficientNet/EfficientNetV2 backbones, indicating that newer is not
  necessarily better for small-data frozen transfer on this task.
  \item \textbf{A calibrated clinical-grade evaluation.} We provide conformal
  prediction intervals, a method-vs-reference Bland--Altman analysis stated with
  its single-reference caveat, rotation test-time augmentation, and Grad-CAM
  attribution, plus a classical structure-tensor prior whose circular fusion with
  the learned model reaches \SI{2.72}{\degree} MAE.
\end{itemize}

%% file: sections/methods.tex
\section{Methods}
\label{sec:methods}

We report a single estimator under two sampling protocols. This section proceeds in the order of the pipeline. We first describe the dataset and the rotation-augmentation protocol that supplies the angle labels (\S\ref{sec:methods:data}). We then define the two sampling protocols and state what each one measures (\S\ref{sec:methods:protocols}). Next come the frozen-backbone feature extractor and the orientation-preserving grid-pooling head suited to an angular target (\S\ref{sec:methods:backbone}), the regression head and its Optuna tuning on cached features (\S\ref{sec:methods:head}), and the cross-validated ensembles that give our best estimator (\S\ref{sec:methods:ensemble}). Finally we describe a classical structure-tensor prior and its circular fusion with the learned estimate (\S\ref{sec:methods:fusion}) and the evaluation suite (\S\ref{sec:methods:eval}). Figure~\ref{fig:pipeline} summarizes the full data-to-evaluation pipeline.

\input{figures/pipeline}

\subsection{Data and rotation augmentation}
\label{sec:methods:data}

The cohort is the one introduced in the original EMBC~2019 study \citep{Patil2019}, drawn from the public SPLab (Brno) ultrasound database. It comprises \num{84} B-mode images of the common carotid artery in longitudinal section, from ten volunteers (mean age $27.5\pm3.5$ years, usually eight images per volunteer). The subjects are young and healthy, the site is the right common carotid in longitudinal view, and acquisition used a single scanner; we draw no conclusions about diseased, aged, or tortuous vessels from this cohort. Images were acquired on a Sonix~OP scanner with two linear-array transducers (\num{10} and \SI{14}{\mega\hertz}, chosen for superficial vascular scanning). The acquisition protocol was fixed: subject supine, neck rotated to the left, right common carotid imaged at a resolution of roughly $390\times330$ pixels. The supervised target is the Doppler angle $\theta$, the angle between the ultrasound beam and the long axis of the vessel (Fig.~\ref{fig:angle_convention}), which enters the Doppler equation~\eqref{eq:doppler} through $\cos\theta$ and is set by hand at the scanner. Because the recovered velocity scales as $1/\cos\theta$, an error in $\theta$ is a leading, operator-dependent source of velocity error (\S\ref{sec:intro}). The database carries no clinical Doppler-angle annotation, so a reference $\theta$ was measured offline by one operator in a custom MATLAB interface, drawing a line along the vessel wall and recording its inclination to the image's vertical axis, with $\theta\in[0,\SI{180}{\degree}]$. This single hand-drawn reading is the only ground truth available per image; its own measurement error is unknown. The estimator reads $\theta$ directly from a single grayscale image, using neither color Doppler nor vessel segmentation.

\input{figures/angle_convention}

\begin{figure}[htbp]
  \centering
  \includegraphics[width=0.86\linewidth]{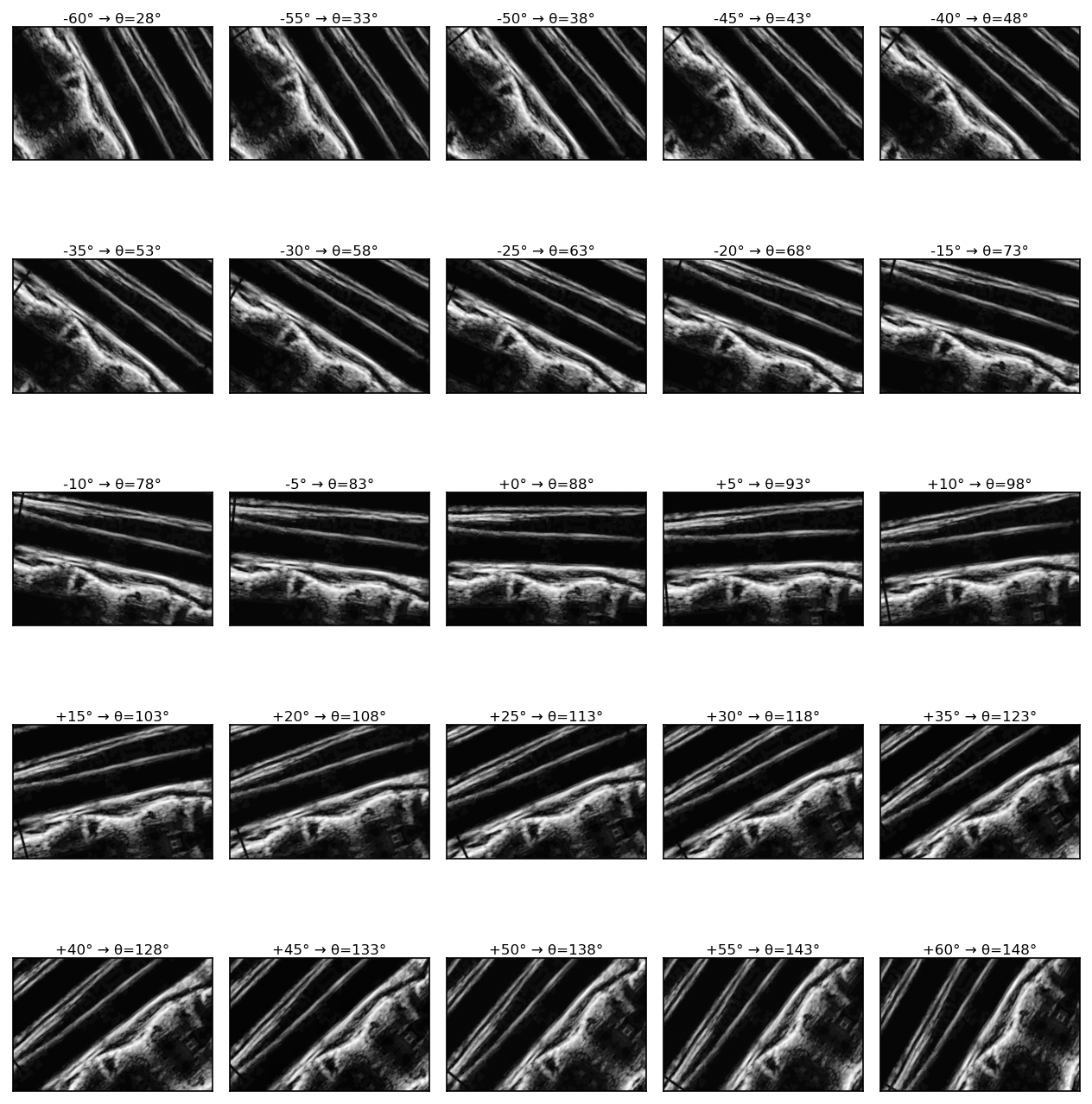}
  \caption{Rotation-augmentation sweep for a single base image: the image is rotated through $[-60,+60]\si{\degree}$ in \SI{5}{\degree} steps (\num{25} views), and the applied rotation defines each view's exact Doppler-angle label (panel titles). CLAHE contrast equalization is applied before feature extraction. This manufactures a labelled regression corpus of $\num{84}\times\num{25}=\num{2100}$ views from the \num{84} base images.}
  \label{fig:augmentation}
\end{figure}

To build a regression corpus with known labels, each base image is rotated through a fixed grid of angles spanning $[-\num{60},+\num{60}]\,\si{\degree}$ in \SI{5}{\degree} steps, yielding \num{25} oriented views per image and $\num{84}\times\num{25}=\num{2100}$ augmented views in total. The label of a rotated view is the base-image reading plus the applied rotation. The applied increment is exact, so the network learns to read off relative image orientation; the label is therefore exact only relative to the single base-image reading, and inherits whatever error that reading carries. Prior to feature extraction each view is contrast-equalized with Contrast-Limited Adaptive Histogram Equalization (CLAHE) \citep{Zuiderveld1994}, which normalizes the wide dynamic range of B-mode speckle and stabilizes the frozen-backbone features across acquisitions. This rotation-augmentation recipe is a standard way to build a labelled regression corpus from a small set of base images. The \num{2100} views are not \num{2100} independent observations: they are \num{25} geometrically coupled copies of each of \num{84} base images, drawn from ten volunteers, giving a three-level hierarchy (volunteer $\rightarrow$ base image $\rightarrow$ rotated view) that the two sampling protocols below treat differently. Figure~\ref{fig:augmentation} shows the rotation sweep for one base image, each view labelled with its resulting angle.

\subsection{Two sampling protocols}
\label{sec:methods:protocols}

The hierarchical structure of the corpus admits more than one principled train/test partition. We report and tune the estimator under two complementary sampling protocols, each answering a distinct question:

\begin{description}[leftmargin=1.25em,style=nextline]
  \item[Image-level sampling :] The \num{2100} rotated views are partitioned at random over the augmented corpus. This is the original study's protocol. It measures angle accuracy across the full population of orientations and imaging conditions the augmentation spans, the regime in which the published headline numbers were reported. Views of one base image can fall on both sides of the split, so this protocol does not test transfer to a new subject.
  \item[Patient-level sampling :] Whole subjects are held out with a grouped split (\textsc{GroupKFold}), so neither a test subject's rotated views nor any of their other images appear in training. The SPLab filenames do not encode volunteer identity, so we recover subject groups by clustering on acquisition time. This yields \num{12} patient-proxy groups over the \num{84} images, more than the ten volunteers, so the proxy grouping is a heuristic and we have no ground-truth subject labels to validate it against; the patient-disjoint guarantee is only as good as that clustering. These \num{12} groups are the unit of all patient-level cross-validation and per-patient aggregation in this paper. The groups are markedly imbalanced (from a single base image up to \num{21}, i.e.\ \num{25}--\num{525} rotated views), so one dominant group can land in a single fold. This inflates per-fold variance and is reflected in the large patient-level standard deviations. The protocol is a stricter lens: it measures generalization to a previously unseen patient. That is harder, because the estimator can no longer rely on subject-specific appearance.
\end{description}

The two protocols answer different questions, and we report both. Holding the model, backbone, head, and training budget fixed while changing only the sampling protocol gives an informative measurement: the patient-level figure quantifies how much of the within-population accuracy is anatomy-specific. Across backbones the patient-level MAPE is roughly double the image-level MAPE, which is what one expects from a small, single-center cohort. The patient-level numbers are the clinically relevant ones, and we read them as such. We tune each protocol separately and report both throughout.

\subsection{Frozen backbones and the grid-pooling head}
\label{sec:methods:backbone}

We use ImageNet-pretrained \citep{Deng2009} convolutional backbones as \emph{frozen} feature extractors, with the classification top removed (\texttt{include\_top=False}); no backbone weights are updated. We evaluate five architectures from the classical ImageNet generation: VGG19 \citep{Simonyan2015}, ResNet50 \citep{He2016}, DenseNet201 \citep{Huang2017}, Xception \citep{Chollet2017}, and InceptionV3 \citep{Szegedy2016}. Freezing keeps the comparison faithful to the compute regime of the original work. It also makes the feature maps cacheable, so the head search of \S\ref{sec:methods:head} runs on precomputed vectors. Figure~\ref{fig:architecture} shows the architecture.

\input{figures/architecture}

The main design choice is how the final spatial feature map is collapsed into a fixed-length vector. The conventional global average pooling (GAP) head averages activations over all spatial locations and is, by construction, approximately rotation-invariant, so it discards the orientation signal the target depends on. That is the wrong inductive bias for an angular target. We replace GAP with an orientation-preserving grid-pooling head: the final feature map is average-pooled onto a small $g\times g$ spatial grid and the per-cell descriptors are flattened and concatenated, so the spatial location of each activation is retained. The frozen backbone already encodes the vessel orientation, and the work is in not pooling it away.

Restoring this coarse spatial structure roughly halves the error. Under image-level five-fold cross-validation, frozen DenseNet201 improves from \SI{10.85}{\percent} MAPE with a GAP head to \SI{4.58}{\percent} with grid pooling. On the original single augmented split it reaches \SI{5.84}{\percent} MAPE (\SI{3.77}{\degree} MAE, $R^2=0.982$), which is the protocol and the number reported in Table~I of the EMBC~2019 paper; this single-split figure is the one that replicates the published result, while the \SI{4.58}{\percent} cross-validated figure is a different estimate under a different protocol. Table~\ref{tab:replication} reports the replication across all five frozen backbones, with DenseNet201 the strongest.

A frozen bake-off under patient-level five-fold cross-validation supports the backbone choice, with the caveat that it rests on \num{84} base images and the same large fold variance noted above, so the ordering should be read against the reported standard deviations rather than as a clean ranking. On the extraction-matched comparison, frozen DenseNet201 (\SI{14.1}{\percent} MAPE) edges out the best modern backbone, ConvNeXt-Base \citep{Liu2022} (\SI{15.7}{\percent} MAPE), with the EfficientNet \citep{Tan2019} family at $\sim\!17$--\SI{21}{\percent} MAPE (Sec.~\ref{subsec:bakeoff}). On this narrow domain, no newer or larger backbone clearly improves on the 2017 DenseNet. DenseNet201, which gives the best $R^2$, wins the replication, and remains strongest after tuning, is carried forward as the primary backbone.

\subsection{Regression head and Optuna tuning on cached features}
\label{sec:methods:head}

On top of the grid-pooled features we place a compact head, \texttt{BatchNorm} $\rightarrow$ \texttt{Dense} $\rightarrow$ \texttt{ReLU} $\rightarrow$ \texttt{Dropout} $\rightarrow$ \texttt{Dense}, that outputs the scalar angle estimate. All models are trained with Adam \citep{Kingma2015} at base learning rate \num{1e-4} under a mean-squared-error loss; the backbone stays frozen throughout. Because the frozen, grid-pooled features can be computed once and cached, head training and the entire hyperparameter search operate on these cached vectors, which makes a large search affordable without a GPU.

We tune the head with the Tree-structured Parzen Estimator (TPE) sampler \citep{Bergstra2011} as implemented in Optuna \citep{Akiba2019}, searching over head depth and width, dropout rate, $L_2$ weight decay, whether batch normalization is applied, learning rate, batch size, and early-stopping patience. Each protocol is tuned separately: every objective evaluation uses the same sampling protocol it reports, scored on that protocol's five-fold cross-validation, so the image-level and patient-level estimators are each optimized for the question they answer. A single feature extraction per backbone serves both protocols. Because the search selects hyperparameters on the same five folds that are then reported, the tuned single-model figures are optimistic relative to a held-out set: they are in-sample to the search, as no nested or outer test fold is used. Tuning moves the single-model DenseNet201 result, with frozen and tuned both scored by five-fold cross-validation (mean over folds), from \SI{4.58}{\percent} to \SI{4.03}{\percent} MAPE (\SI{3.00}{\degree} MAE) under image-level sampling, and from \SI{12.59}{\percent} to \SI{10.80}{\percent} MAPE (\SI{8.62}{\degree} MAE, $R^2=0.886$) under patient-level sampling. The per-model gains are small but consistent, and they calibrate the five members enough that a plain mean ensemble becomes effective.

\subsection{Cross-validated ensembles}
\label{sec:methods:ensemble}

We combine the tuned heads over all five backbones. Two combiners are reported: a plain mean of the member predictions, and a stacked generalization \citep{Wolpert1992} in which a Ridge meta-learner is fit on the member outputs. The stacker and every reported ensemble figure are computed from out-of-fold (OOF) predictions, pooled across folds into a single metric: under five-fold cross-validation each member predicts only on the fold it did not train on, so under patient-level sampling every ensemble number is evaluated on patients absent from the corresponding fold's training. This pooled-OOF aggregation differs from the mean-over-folds aggregation used for the single-model figures of \S\ref{sec:methods:head}, so single-model and ensemble metrics are not strictly the same estimator and should be compared with that in mind. The Ridge meta-learner is fit on OOF member outputs over only \num{12} patient groups, which leaves the meta-layer at some risk of overfitting; we did not run a separate nested OOF loop for the stacker. The mean ensemble reaches \SI{3.03}{\percent} MAPE (\SI{2.09}{\degree} MAE, $R^2=0.994$) under image-level sampling and \SI{9.89}{\percent} MAPE (\SI{6.89}{\degree} MAE, $R^2=0.932$) under patient-level sampling. The stacked ensemble is the estimator we carry forward, reaching \SI{2.79}{\percent} MAPE (\SI{1.96}{\degree} MAE, $R^2=0.995$) under image-level sampling and \SI{8.53}{\percent} MAPE (\SI{5.93}{\degree} MAE, $R^2=0.952$) under patient-level sampling. Ensembling gives the largest single gain we observe: combining the five backbones recovers accuracy that no single model reaches, consistent with the members making partly independent errors, though we do not report an error-correlation analysis to quantify that independence. The patient-level OOF predictions are also the substrate for the calibrated reporting of \S\ref{sec:methods:eval}.

\subsection{Classical structure-tensor prior and circular fusion}
\label{sec:methods:fusion}

To situate the learned estimator against a transparent geometric baseline, we also compute a purely classical, image-only orientation prior from the structure tensor: the dominant local-gradient orientation, aggregated over the vessel region, gives an angle estimate with no learning and no training data. Evaluated on the narrow band of base-image angles, this prior reaches \SI{3.16}{\degree} MAE. Because an orientation is defined only up to a half-turn, we fuse the learned and classical estimates in the circular $2\theta$ domain, averaging $(\cos 2\theta,\sin 2\theta)$ before mapping back. On the same base-image set the fused estimate reaches \SI{2.72}{\degree} MAE, below either component alone. This comparison is on \num{84} base images and a different, smaller evaluation set than the rotation-augmented corpus used elsewhere; we read the fused gain as suggestive of partly complementary information between the learned model and the geometric cue, not as an established decomposition, since we report no residual correlations or confidence intervals.

\subsection{Evaluation suite}
\label{sec:methods:eval}

Beyond a point estimate, we report calibrated diagnostics, all derived from the patient-level OOF predictions of the tuned DenseNet201. We construct distribution-free, finite-sample prediction intervals by split conformal regression \citep{Vovk2005,Lei2018}, using a patient-disjoint calibration/test partition so that calibration residuals come from subjects absent from both the model's training and the reported test set. Marginal conformal coverage is a guarantee in expectation; on a cohort of \num{12} imbalanced groups the calibration set is small, so the coverage actually observed on the single test partition is itself noisy, and the achieved coverage and calibration-set size should be read alongside the interval width in \S\ref{sec:evaluation}. Agreement with the reference angle is summarized by a Bland--Altman analysis \citep{BlandAltman1986}, reporting bias and limits of agreement. Because only a single reference reading (the MATLAB-GUI angle) is available per image, this is a method-versus-reference comparison, not an inter-observer study, and the precision of the reference is unknown. At inference we apply rotation test-time augmentation, predicting each base image over its rotated views and aggregating by the circular median, which reduces base-image MAE; note that this reuses the same rotation family that generated the labels, so the gain is measured over transformations of training-adjacent data rather than independent views. Finally, Grad-CAM \citep{Selvaraju2017} attributions are computed on the frozen backbone as a qualitative check that the estimator keys on the vessel wall, the anatomy that defines the flow axis, rather than on background structure; this is a saliency inspection, not a formal verification. The full clinical evaluation is reported in \S\ref{sec:evaluation}.

%% file: figures/pipeline.tex
\begin{figure}[htbp]
\centering
\resizebox{\linewidth}{!}{%
\begin{tikzpicture}[
  font=\small,
  >={Stealth[length=2.2mm]},
  every node/.style={align=center},
  stage/.style={draw, rounded corners=2pt, minimum height=11mm, minimum width=20mm, fill=blue!5},
  data/.style={stage, fill=black!7},
  proto/.style={draw, rounded corners=2pt, minimum height=7mm, minimum width=24mm, fill=orange!14, font=\footnotesize},
  eval/.style={stage, fill=green!8, minimum width=24mm},
  node distance=8mm and 10mm,
]
  \node[data]  (raw)  {84 base\\images};
  \node[stage, right=of raw]  (aug)  {augment\\$\pm60^\circ/5^\circ$ + CLAHE};
  \node[stage, right=of aug]  (corp) {2{,}100-image\\corpus};
  \node[stage, right=of corp] (feat) {cached frozen\\features\\(1 / backbone)};
  \node[stage, right=of feat] (tune) {Optuna TPE\\tuning\\(per protocol)};
  \node[stage, right=of tune] (ens)  {5-backbone\\ensemble\\mean / stacked};
  \node[eval,  right=of ens]  (ev)   {evaluation};

  \draw[->] (raw)--(aug); \draw[->] (aug)--(corp); \draw[->] (corp)--(feat);
  \draw[->] (feat)--(tune); \draw[->] (tune)--(ens); \draw[->] (ens)--(ev);

  \node[proto, above=8mm of tune] (pimg) {image-level (random)};
  \node[proto, below=8mm of tune] (ppat) {patient-level (GroupKFold)};
  \draw[->, dashed, gray] (pimg) -- (tune);
  \draw[->, dashed, gray] (ppat) -- (tune);

  \node[below=7mm of ev, text width=30mm, font=\footnotesize] (evlist)
       {conformal intervals,\\Bland--Altman, TTA,\\Grad-CAM,\\structure-tensor fusion};
  \draw[->, dashed, gray] (ev) -- (evlist);

  \begin{scope}[on background layer]
    \node[draw=black!25, dashed, rounded corners, fit=(pimg)(ppat),
          inner sep=2mm, label=left:{\footnotesize two protocols}] {};
  \end{scope}
\end{tikzpicture}}
\caption{Experiment pipeline. The 84 base images are expanded by a deterministic
rotation sweep (with CLAHE) into a $\sim$2{,}100-image corpus. Each backbone's
frozen features are extracted once and cached, then scored under \emph{two
complementary sampling protocols} - image-level (random) and patient-level
(grouped by volunteer) - with per-protocol Optuna TPE tuning of the head. The
five tuned backbones are combined (mean or stacked ensemble) and assessed with a
clinical-grade evaluation suite.}
\label{fig:pipeline}
\end{figure}

%% file: figures/angle_convention.tex
\begin{figure}[htbp]
\centering
\begin{tikzpicture}[
  font=\small,
  >={Stealth[length=2.4mm]},
  axis/.style={->, blue!55!black, line width=1.0pt},
  vessel/.style={line width=8pt, gray!30, line cap=round},
]
  \draw[gray!40, rounded corners=1pt] (-2.8,-2.5) rectangle (3.0,2.5);
  \node[gray!60, font=\footnotesize] at (1.75,-2.25) {B-mode frame};

  \def\thetadeg{58}
  \draw[vessel] ({-2.4*sin(\thetadeg)},{-2.4*cos(\thetadeg)})
             -- ({2.4*sin(\thetadeg)},{2.4*cos(\thetadeg)});
  \draw[black, line width=0.9pt]
       ({-2.4*sin(\thetadeg)},{-2.4*cos(\thetadeg)})
    -- ({2.4*sin(\thetadeg)},{2.4*cos(\thetadeg)});
  \node[font=\footnotesize, anchor=south west]
       at ({2.0*sin(\thetadeg)},{2.0*cos(\thetadeg)}) {vessel wall};
  \draw[->, red!55!black, line width=0.9pt]
       (0,0) -- ({1.15*sin(\thetadeg)},{1.15*cos(\thetadeg)});
  \node[red!55!black, font=\footnotesize]
       at ({1.15*sin(\thetadeg)+0.35},{1.15*cos(\thetadeg)-0.12}) {flow};

  \draw[axis] (0,2.45) -- (0,-2.45);
  \node[blue!55!black, font=\footnotesize, anchor=south east] at (-0.12,1.9)
       {beam axis};
  \node[blue!55!black, font=\footnotesize, anchor=north east] at (-0.12,1.88)
       {(image vertical)};

  \draw[->, black, line width=0.9pt]
       (0,1.0) arc[start angle=90, end angle={90-\thetadeg}, radius=1.0];
  \node at ({0.62*sin(\thetadeg/2)},{0.62*cos(\thetadeg/2)+0.32}) {$\theta$};
\end{tikzpicture}
\caption{Doppler-angle convention. The Doppler angle $\theta$ is the angle
between the ultrasound beam (the image's vertical axis) and the vessel long axis,
read off a single B-mode frame by aligning a line with the vessel wall and
measuring its inclination to the vertical; $\theta$ ranges over
$[0,\SI{180}{\degree}]$ and is direction-agnostic. The recovered velocity scales
as $1/\cos\theta$, so the accuracy of this angle directly controls the accuracy of
the reported velocity.}
\label{fig:angle_convention}
\end{figure}

%% file: figures/architecture.tex
\begin{figure}[htbp]
\centering
\resizebox{\linewidth}{!}{%
\begin{tikzpicture}[
  font=\small,
  >={Stealth[length=2.2mm]},
  every node/.style={align=center},
  box/.style={draw, rounded corners=2pt, minimum height=11mm, minimum width=17mm, fill=blue!4},
  frozen/.style={box, fill=black!8},
  pool/.style={box, fill=orange!16},
  head/.style={box, fill=green!8},
  io/.style={box, fill=blue!10, minimum width=13mm},
  node distance=7mm,
]
  \node[io]      (img)  {B-mode\\image\\$H{\times}W$};
  \node[frozen, right=of img]  (bb)  {frozen ImageNet\\backbone\\\texttt{include\_top=False}};
  \node[box,    right=of bb]   (fmap){conv map\\$h{\times}w{\times}c$};
  \node[pool,   right=of fmap] (grid){\textbf{grid pool}\\$G{\times}G$};
  \node[box,    right=of grid] (flat){flatten\\$G^2c$};
  \node[head,   right=of flat] (head){BN\,$\to$\,Dense\\$\to$\,ReLU\,$\to$\,Drop\\$\to$\,Dense};
  \node[io,     right=of head] (out) {$\hat\theta$\\(degrees)};

  \draw[->] (img)--(bb); \draw[->] (bb)--(fmap); \draw[->] (fmap)--(grid);
  \draw[->] (grid)--(flat); \draw[->] (flat)--(head); \draw[->] (head)--(out);

  \node[below=5mm of grid, text width=42mm, font=\footnotesize] (note)
       {keeps coarse spatial layout\\($G{\times}G$, orientation-aware)\\[1pt]
        \emph{vs} global avg-pool: $1{\times}1$,\\approximately rotation-invariant};
  \draw[->, dashed, gray] (note) -- (grid);

  \begin{scope}[on background layer]
    \node[draw=black!25, dashed, rounded corners, fit=(bb)(fmap), inner sep=2.5mm,
          label=above:{frozen feature extractor}] {};
    \node[draw=black!25, dashed, rounded corners, fit=(grid)(flat)(head), inner sep=2.5mm,
          label=above:{trained head}] {};
  \end{scope}
\end{tikzpicture}}
\caption{Estimator architecture. A frozen ImageNet backbone produces a
convolutional feature map; \textbf{orientation-preserving grid pooling} averages
that map to a small $G{\times}G$ grid (rather than a single $1{\times}1$ vector)
so the coarse spatial layout that encodes vessel orientation survives. The
flattened grid feeds a shallow regression head that outputs the Doppler angle
$\hat\theta$. Only the head is trained.}
\label{fig:architecture}
\end{figure}

%% file: sections/results.tex
\section{Results}
\label{sec:results}

We report results in four parts that follow the argument of the paper. We first establish a \emph{faithful replication} of the original pipeline under image-level sampling, and isolate the single pooling choice that controls its accuracy (Sec.~\ref{subsec:replication}). We then give the \emph{best estimator under each of the two sampling protocols}, an Optuna-tuned stacked ensemble, and place it against the EMBC~2019 figures (Sec.~\ref{subsec:dual}). We next ask whether a stronger backbone is the right lever, through an \emph{architecture bake-off} under patient-level cross-validation (Sec.~\ref{subsec:bakeoff}). Finally we trace the \emph{methods progression} that yields the patient-level estimator (Sec.~\ref{subsec:progression}). Throughout, $\theta$ denotes the Doppler beam-to-vessel angle, operationalized here as the magnitude of applied image rotation (Sec.~\ref{sec:methods}). Mean absolute error (MAE) and root-mean-square error (RMSE) are reported in degrees, mean absolute percentage error (MAPE) as a percentage, and the coefficient of determination $R^2$ is unitless. All runs use seed~42 under Keras~3 / JAX.

The two sampling protocols are both legitimate and answer different questions. \emph{Image-level sampling} draws the train/test partition over the rotation-augmented corpus and measures accuracy across the population of orientations and imaging conditions the augmentation spans; this is the protocol of the original study. \emph{Patient-level sampling} holds out whole volunteers via \texttt{GroupKFold} and measures generalization to a previously unseen patient. We report and tune both.

\subsection{Faithful replication under image-level sampling}
\label{subsec:replication}

We reproduce the EMBC~2019 pipeline~\citep{Patil2019} under image-level sampling. The 84 longitudinal common-carotid B-mode images are rotation-augmented over $[-60,+60]\si{\degree}$ in \SI{5}{\degree} steps, giving \num{25} oriented views each and $\approx\!2{,}100$ labelled examples, and partitioned into train and test at the image level. Table~\ref{tab:replication} pairs frozen ImageNet backbones~\citep{Deng2009} (\texttt{include\_top=False}, no fine-tuning) with the orientation-preserving grid-pooling head.

The pooling operator is the design choice that governs accuracy here. Global average pooling collapses the convolutional feature map over all spatial positions and is therefore approximately rotation-invariant. That is the wrong inductive bias when the regression target is itself an orientation, because $\theta$ is exactly the quantity a rotation-invariant summary discards. A grid-pooling head instead average-pools the feature map to a small $G\times G$ grid and then flattens, preserving the coarse spatial arrangement of activations and retaining the orientation signal the target depends on.

This change alone roughly halves the frozen DenseNet201~\citep{Huang2017} error. Under image-level five-fold cross-validation it falls from \SI{10.85}{\percent} MAPE with a global-average-pooling head to \SI{4.58}{\percent} with grid pooling. On the original single augmented split (Table~\ref{tab:replication}) the grid-pooling model reaches \SI{5.84}{\percent} MAPE (\SI{3.77}{\degree} MAE, \SI{4.75}{\degree} RMSE, $R^2=0.982$). With this head, DenseNet201 is the strongest backbone by a wide margin, recovering the single-digit-degree error band of the original report under its stated protocol and with no fine-tuning. The other backbones trail well behind under the identical head: InceptionV3~\citep{Szegedy2016} at \SI{11.70}{\percent}, Xception~\citep{Chollet2017} at \SI{12.00}{\percent}, ResNet50~\citep{He2016} at \SI{12.75}{\percent}, and VGG19~\citep{Simonyan2015} at \SI{18.01}{\percent}. This is consistent with DenseNet's dense feature reuse suiting a fine-grained orientation task.

\begin{figure}[htbp]
  \centering
  \includegraphics[width=0.92\linewidth]{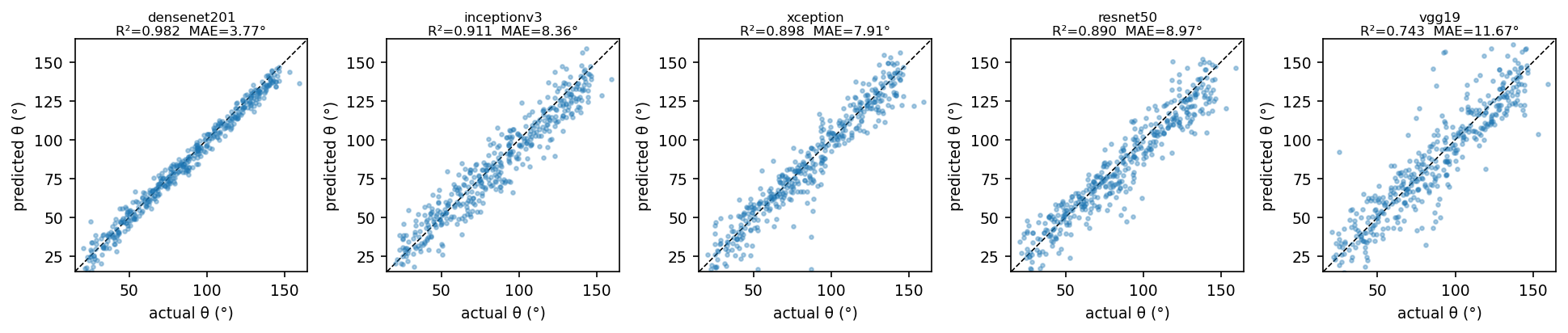}
  \caption{Predicted versus actual Doppler angle for each frozen backbone with the grid-pooling head, under image-level sampling. Points on the diagonal are exact predictions; DenseNet201 tracks the identity line across the full angle range, while the weaker backbones scatter more widely.}
  \label{fig:pred_vs_actual}
\end{figure}

\begin{figure}[htbp]
  \centering
  \includegraphics[width=0.7\linewidth]{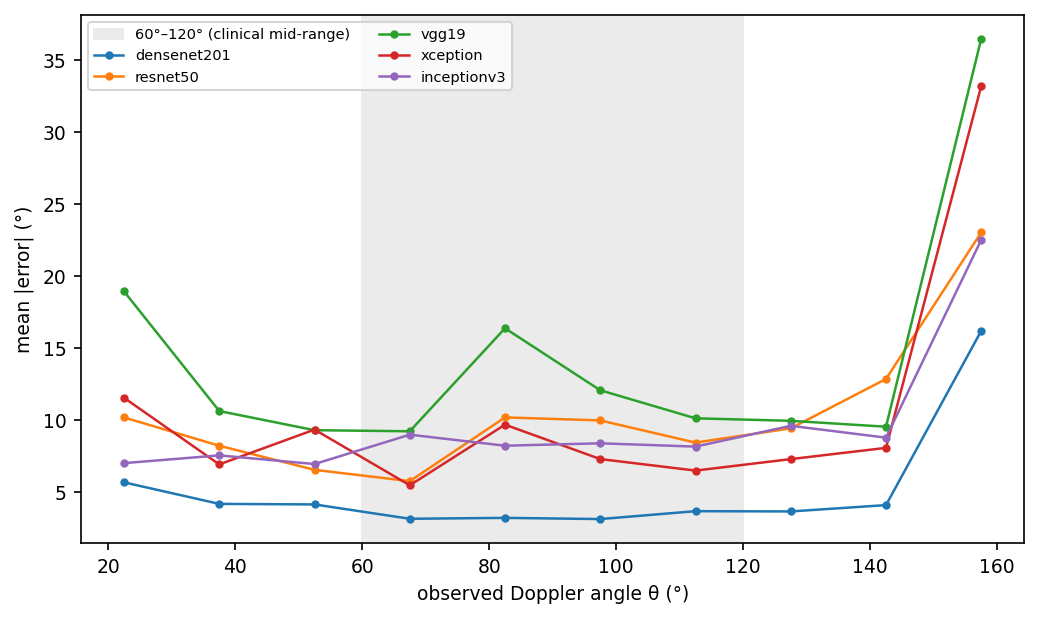}
  \caption{Mean absolute error as a function of the observed Doppler angle. For the best model (DenseNet201) the error is low and roughly flat across the mid-range and rises sharply only at the most oblique observed angle ($\sim$\SI{158}{\degree}); the weaker backbones carry larger and less stable error, with a local rise near \SI{82}{\degree}. The shaded band marks the clinically common $60$--$120\si{\degree}$ range.}
  \label{fig:error_vs_angle}
\end{figure}

One difference from the original report is substantive. Patil and Anand found \emph{VGG19} to be their single best backbone (\SI{2.87}{\degree} MAE), whereas under our shared grid-pooling head and fixed training budget DenseNet201 leads and VGG19 trails the field. We therefore reproduce the original error \emph{regime}, a frozen backbone reaching single-digit-degree error, but not its per-model ranking. The most likely cause is methodological. To keep the comparison controlled we hold the pooling operator, head architecture, and training budget identical across all five backbones, whereas the original tuned these per model. A configuration that suits VGG19's coarse, large-stride feature map need not be the one that suits DenseNet201's dense maps, so a single shared head can reshuffle the ranking even when the attainable accuracy is preserved. Figure~\ref{fig:pred_vs_actual} shows the per-backbone predicted-versus-actual fit, and Fig.~\ref{fig:error_vs_angle} the absolute error across the observed-angle range.

\input{tables/replication}

\subsection{Best estimator under each sampling protocol}
\label{subsec:dual}

Table~\ref{tab:dual_protocol} reports the Optuna-tuned estimator under both sampling protocols, each tuned to its own best configuration. The head and optimizer hyperparameters (head depth and width, dropout, $L_2$ penalty, BatchNorm placement, learning rate, batch size, and early-stopping patience) are searched with the Tree-structured Parzen Estimator~\citep{Bergstra2011} as implemented in Optuna~\citep{Akiba2019}, scored by five-fold cross-validation on cached frozen features. The backbone extraction is run once per network and the cached features serve both protocols, so the two halves of the table are like-for-like up to the sampling level. Per-backbone rows are five-fold cross-validated tuned heads; ensemble rows are out-of-fold (OOF) predictions over the five tuned fold models. The two row types are thus different estimators of the same quantity, a point we return to below. Optimization uses Adam~\citep{Kingma2015} with a base learning rate of \num{1e-4} and an MSE objective; CLAHE~\citep{Zuiderveld1994} contrast normalization is applied to the inputs.

The single best backbone is DenseNet201, which reaches \SI{4.03}{\percent} MAPE (\SI{3.00}{\degree} MAE, $R^2=0.988$) under image-level sampling and \SI{10.80}{\percent} MAPE (\SI{8.62}{\degree} MAE, $R^2=0.886$) under patient-level five-fold cross-validation. Tuning the head moves DenseNet201 from the frozen \SI{4.58}{\percent} to \SI{4.03}{\percent} at image level and from \SI{12.59}{\percent} to \SI{10.80}{\percent} at patient level (frozen and tuned both five-fold CV). These shifts are small relative to the per-fold standard deviations (Table~\ref{tab:dual_protocol}), so we read tuning as a consistent but modest refinement rather than a step change. The per-backbone ordering by MAE is stable across both protocols (DenseNet201 $\geq$ ResNet50 $\geq$ VGG19 $\geq$ InceptionV3 $\approx$ Xception), so the choice of backbone is largely insensitive to the evaluation lens. Under patient-level MAPE, VGG19 at \SI{14.97}{\percent} and InceptionV3 at \SI{14.66}{\percent} sit within \SI{0.3}{} percentage points and exchange ranks.

Ensembling the five tuned backbones improves both columns. A plain OOF mean of the five members reaches \SI{3.03}{\percent} MAPE (\SI{2.09}{\degree} MAE, $R^2=0.994$) under image-level sampling and \SI{9.89}{\percent} MAPE (\SI{6.89}{\degree} MAE, $R^2=0.932$) under patient-level sampling. A stacked combiner~\citep{Wolpert1992} fitted over the OOF predictions sharpens both further, to \textbf{\SI{2.79}{\percent} MAPE (\SI{1.96}{\degree} MAE, $R^2=0.995$)} under image-level sampling and \textbf{\SI{8.53}{\percent} MAPE (\SI{5.93}{\degree} MAE, $R^2=0.952$)} under patient-level sampling. This is the first patient-level result in our progression to fall below \SI{10}{\percent} MAPE.

The size of this gain depends on how the single-model baseline is aggregated, and the like-for-like comparison is the fair one. Aggregating the single tuned DenseNet201 the same pooled-OOF way reaches \SI{7.80}{\degree} MAE / \SI{10.14}{\percent} MAPE at patient level, against its \SI{8.62}{\degree} per-fold mean in Table~\ref{tab:dual_protocol}. On the matched pooled-OOF basis the stacked ensemble improves the single best model by about \SI{1.9}{\degree} MAE. The larger apparent gap (\SI{8.62}{\degree} to \SI{5.93}{\degree}) partly reflects the per-fold-mean versus pooled-OOF aggregation, not ensembling alone, so we quote the matched figure as the ensembling gain. That the plain mean is already competitive follows from tuning: per-member calibration is good enough that a simple average works, whereas averaging the untuned heads, whose error scales differ, was far less effective.

\input{tables/dual_protocol}

\begin{figure}[htbp]
  \centering
  \includegraphics[width=0.92\linewidth]{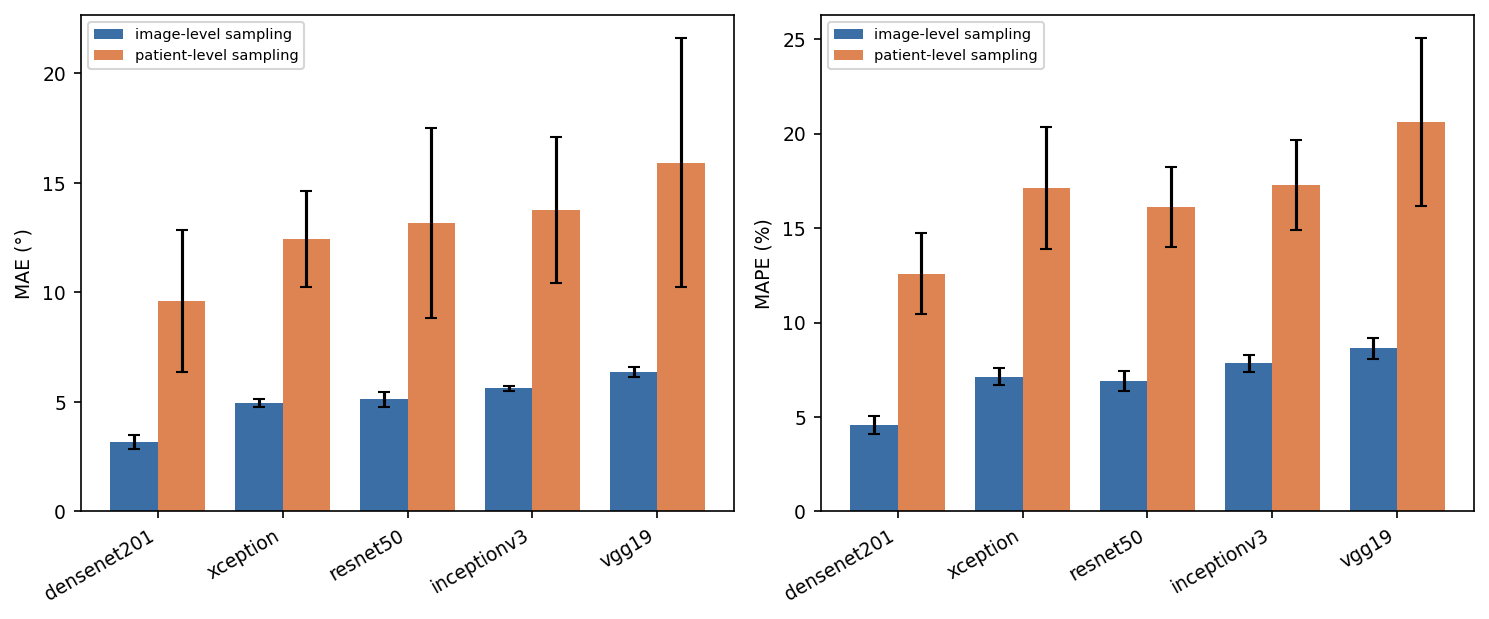}
  \caption{Sampling-protocol sensitivity of the \emph{frozen} backbones with untuned grid-pooling heads (distinct from the Optuna-tuned estimator of Table~\ref{tab:dual_protocol}): per-backbone MAE (left) and MAPE (right) under image-level versus patient-level sampling. Both protocols are reported as five-fold cross-validation means with $\pm$1 standard-deviation whiskers. The cross-subject (patient-level) protocol is the harder regime, scoring generalization to unseen anatomy rather than across the augmented-image population, so every backbone shows a larger and more variable error there.}
  \label{fig:protocol_gap}
\end{figure}

The same estimator reports a higher error under patient-level sampling than under image-level sampling: for the stacked ensemble, \SI{8.53}{\percent} against \SI{2.79}{\percent} MAPE, roughly a threefold ratio. This is the expected signature of a small cohort. We recover \num{12} patient-proxy groups from acquisition-time clustering rather than from ground-truth subject identifiers (Sec.~\ref{sec:methods}), so every patient-level number, including this one, rests on a reconstructed grouping. The image-level fit is tight because the model is evaluated within the augmented-image population it was trained on; cross-subject generalization is intrinsically harder and carries real per-fold variance, with few held-out groups per fold and groups that range widely in size. The patient-level point values in this paragraph are pooled-OOF and therefore have no per-fold whiskers; the per-fold spread that does exist is reported in the table caption, and it is large enough that the patient-level numbers should be read as estimates from a small, unequal split, not as tight central values. The image-level versus patient-level gap quantifies how much of the within-population accuracy is anatomy-specific. Neither protocol is defective; they measure different things. A reader interested in generalization to a new patient should weigh the patient-level number, and a reader interested in accuracy across the imaging conditions the corpus spans should weigh the image-level number. Figure~\ref{fig:protocol_gap} shows the same protocol sensitivity for the \emph{frozen} backbones, backbone by backbone; the tuned ensemble shows an analogous, somewhat smaller gap.

\paragraph{Comparison to EMBC~2019.} The best single model of the original study reports $\approx\!\SI{2.87}{\degree}$ MAE / \SI{4.03}{\percent} MAPE under image-level sampling~\citep{Patil2019}. This work recovers that error band with a single tuned DenseNet201 (\SI{3.00}{\degree} MAE / \SI{4.03}{\percent} MAPE) and improves on it with the tuned stacked ensemble (\SI{1.96}{\degree} MAE / \SI{2.79}{\percent} MAPE, $R^2=0.995$), while additionally reporting the previously unmeasured patient-level number. The MAPE figures coincide at \SI{4.03}{\percent}; this is a numerical coincidence of two independently computed quantities, not a fitted match, and the MAE differs ($\SI{3.00}{\degree}$ here versus $\SI{2.87}{\degree}$ originally).

\subsection{Architecture bake-off: newer is not better}
\label{subsec:bakeoff}

\begin{figure}[htbp]
  \centering
  \includegraphics[width=0.78\linewidth]{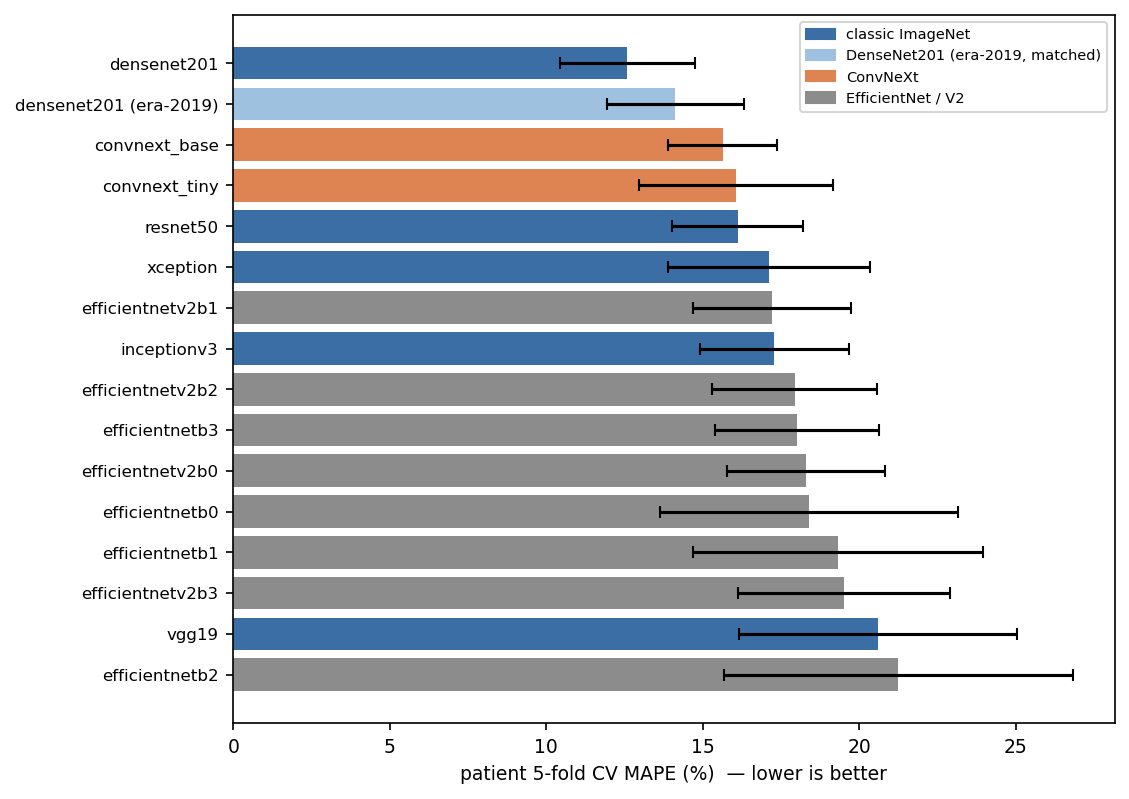}
  \caption{Architecture bake-off under patient-level five-fold cross-validation (lower is better; bars are fold mean, whiskers $\pm$ one fold standard deviation; colour denotes the backbone family). All encoders are frozen with the identical grid-pooling head. A frozen DenseNet201 is the single best backbone, ahead of every modern ConvNeXt and EfficientNet / EfficientNetV2 variant; ConvNeXt is competitive with the other classic networks, so the result is that no newer or larger encoder improved on the 2017 DenseNet201 on this small, out-of-domain task, not that classic architectures win as a family. The five classic bars are the replication-era frozen extraction (\texttt{cmp\_*} rows); the ConvNeXt/EfficientNet bars are the era-2019 extraction (\texttt{f\_*}/\texttt{e2\_*} rows). The ordering is robust to this: even DenseNet201's era-2019 number (\SI{14.1}{\percent}) leads ConvNeXt-Base (\SI{15.7}{\percent}).}
  \label{fig:bakeoff}
\end{figure}

A natural way to improve the patient-level estimator is to reach for a stronger or more recent backbone. We run an architecture bake-off under patient-level five-fold cross-validation, holding the grid-pooling head and training budget fixed and varying only the frozen encoder across classic ImageNet networks and the modern ConvNeXt~\citep{Liu2022} and EfficientNet / EfficientNetV2~\citep{Tan2019} families. Fig.~\ref{fig:bakeoff} summarizes the comparison.

No newer or larger backbone improves on DenseNet201. On the same era-2019 extraction that produced the modern bars, frozen DenseNet201 scores \SI{14.13}{\percent} ($\pm$\SI{2.19}{\percent}) MAPE, ahead of the best modern backbone, ConvNeXt-Base, at \SI{15.65}{\percent} ($\pm$\SI{1.75}{\percent}). These one-sigma bands overlap, so the margin is a consistent ordering across folds rather than a separation we can call significant; we report it as the matched comparison, not as a decisive gap. ConvNeXt-Tiny follows at \SI{16.07}{\percent}, and the EfficientNet / EfficientNetV2 B0--B3 variants trail from \SI{17}{\percent} to \SI{21.25}{\percent}. In its own fresh replication-era extraction DenseNet201 is lower still, \SI{12.59}{\percent}, but we lead with the matched number so the comparison is like-for-like.

The remaining classic backbones were extracted in the replication-era pipeline (ResNet50 \SI{16.12}{\percent}, Xception \SI{17.11}{\percent}, InceptionV3 \SI{17.29}{\percent}, VGG19 \SI{20.60}{\percent}), so they cannot be ranked against ConvNeXt strictly like-for-like, and we do not claim a firm classic-versus-ConvNeXt ordering among the non-DenseNet networks. Only DenseNet201 has a number on both extraction pipelines; the rest of the ranking is confounded by extraction era and should be read with that caveat. What is robust to extraction era is the narrow claim we make: on \num{84} base images of one narrow domain, with frozen features and a fixed head, no backbone newer or larger than the 2017 DenseNet201 beat it. We do not attribute this to a capacity ceiling, since we have not measured representation quality across encoders; the observation is bounded by this dataset and this protocol. DenseNet201 is the replication winner, the best frozen result here, and the strongest after tuning, so we carry it forward. This motivates the complementary strategy of the previous subsection: rather than change the backbone, tune the head and ensemble.

\subsection{Methods progression under patient-level sampling}
\label{subsec:progression}

\begin{figure}[htbp]
  \centering
  \includegraphics[width=0.62\linewidth]{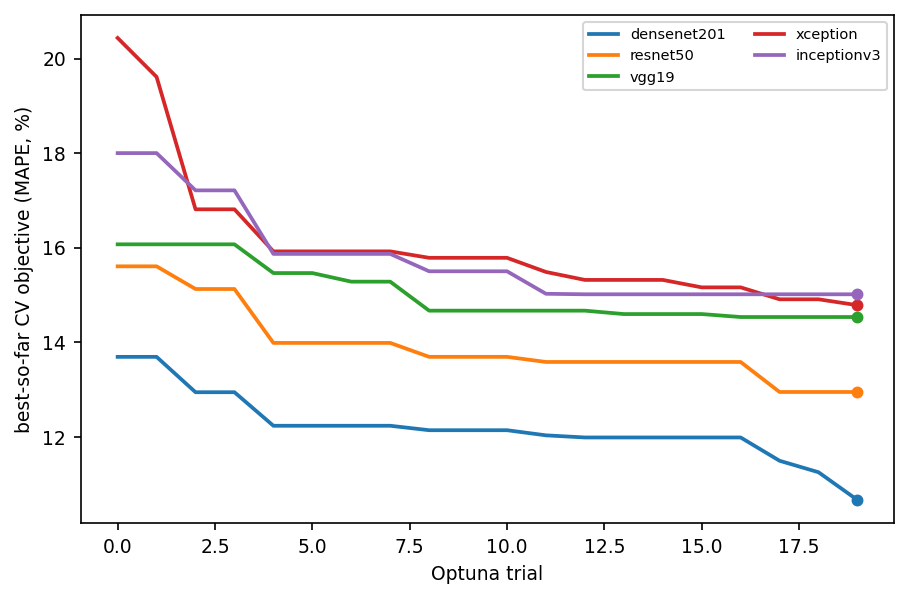}
  \caption{Optuna TPE optimization history (best-so-far cross-validation objective vs trial) for each backbone's grid-pooling head and training hyperparameters under patient-level five-fold cross-validation. For DenseNet201 the best trial reaches a \SI{10.67}{\percent} search objective; the selected configuration, re-evaluated, scores \SI{10.80}{\percent} per-fold (Table~\ref{tab:methods_progression}), before ensembling.}
  \label{fig:tuning}
\end{figure}

Table~\ref{tab:methods_progression} traces how the patient-level estimator is built from frozen DenseNet201, with every step under the same patient five-fold cross-validation. The orientation-preserving grid-pooling head cuts the global-average-pooling baseline from \SI{18.70}{\percent} ($\pm$\SI{3.24}{\percent}) to \SI{12.59}{\percent} ($\pm$\SI{2.15}{\percent}) MAPE. Tuning that head with the Optuna TPE sampler~\citep{Akiba2019,Bergstra2011} brings it to \SI{10.80}{\percent} ($\pm$\SI{2.59}{\percent}), and Fig.~\ref{fig:tuning} shows the corresponding optimization history. The largest single gain comes from ensembling: combining the five tuned, patient-disjoint fold models and reading out their OOF predictions, a simple mean reaches \SI{9.89}{\percent} MAPE and a stacked combiner~\citep{Wolpert1992} reaches \SI{8.53}{\percent}, less than half the GAP baseline, entirely within the patient-level protocol.

Every row is measured under the same patient five-fold cross-validation, so the steps are directly comparable. The clinical-grade evaluation of the patient-level tuned ensemble is reported in Sec.~\ref{sec:evaluation}: conformal prediction intervals~\citep{Vovk2005,Lei2018}, Bland--Altman agreement~\citep{BlandAltman1986} against the single reference reading, test-time augmentation, and Grad-CAM~\citep{Selvaraju2017} attribution.

\input{tables/methods_progression}

%% file: tables/replication.tex
\begin{table}[htbp]
  \centering
  \caption{Faithful replication under image-level sampling (the original augmented-corpus protocol): frozen ImageNet backbones with the orientation-preserving grid-pooling head. The best configuration (DenseNet201) reaches \SI{5.84}{\percent} MAPE, reproducing Table~I of the EMBC 2019 paper.}
  \label{tab:replication}
  \begin{tabular}{llrrrrr}
    \toprule
    Backbone & Pooling & $N_{\mathrm{test}}$ & MAE (\degr) & RMSE (\degr) & MAPE (\%) & $R^2$ \\
    \midrule
    DenseNet201 & grid3 & 420 & 3.77 & 4.75 & 5.84 & 0.982 \\
    InceptionV3 & grid2 & 420 & 8.36 & 10.44 & 11.70 & 0.911 \\
    Xception & grid2 & 420 & 7.91 & 11.20 & 12.00 & 0.898 \\
    ResNet50 & grid2 & 420 & 8.97 & 11.62 & 12.75 & 0.890 \\
    VGG19 & grid2 & 420 & 11.67 & 17.77 & 18.01 & 0.743 \\
    \bottomrule
  \end{tabular}
\end{table}

%% file: tables/dual_protocol.tex
\begin{table}[htbp]
  \centering
  \caption{Doppler-angle estimation under two sampling protocols, each Optuna-tuned to its own best configuration. \emph{Image-level sampling} draws the train/test partition over the rotation-augmented image corpus (the standard synthetic-data protocol of the original study); \emph{patient-level sampling} holds out whole volunteers, a stricter cross-subject generalization test. Per-backbone rows are patient/image five-fold cross-validated tuned heads (means; per-fold MAPE std is \SIrange{0.1}{0.7}{\percent} for image-level and \SIrange{2.6}{4.4}{\percent} for the higher-variance patient-level folds); ensemble rows are pooled out-of-fold over the five tuned models. Note the aggregation differs: member rows are per-fold means, ensemble rows are pooled OOF (the single tuned DenseNet201 aggregated pooled-OOF is \SI{7.80}{\degree} MAE patient-level, the like-for-like predecessor of the ensemble rows). Both columns are recomputed from \texttt{results/} by \texttt{scripts/gen\_paper\_tables.py}.}
  \label{tab:dual_protocol}
  \begin{tabular}{lrrrrrr}
    \toprule
    & \multicolumn{3}{c}{Image-level sampling} & \multicolumn{3}{c}{Patient-level sampling} \\
    \cmidrule(lr){2-4}\cmidrule(lr){5-7}
    Model (Optuna-tuned) & MAE (\degr) & MAPE (\%) & $R^2$ & MAE (\degr) & MAPE (\%) & $R^2$ \\
    \midrule
    DenseNet201 & 3.00 & 4.03 & 0.988 & 8.62 & 10.80 & 0.886 \\
    ResNet50 & 3.64 & 4.84 & 0.981 & 9.79 & 13.30 & 0.842 \\
    VGG19 & 4.22 & 5.79 & 0.976 & 10.21 & 14.97 & 0.871 \\
    InceptionV3 & 4.68 & 6.59 & 0.970 & 10.66 & 14.66 & 0.856 \\
    Xception & 4.75 & 6.59 & 0.970 & 10.88 & 14.76 & 0.851 \\
    \midrule
    5-model ensemble (mean) & 2.09 & 3.03 & 0.994 & 6.89 & 9.89 & 0.932 \\
    \textbf{5-model ensemble (stacked)} & 1.96 & 2.79 & 0.995 & 5.93 & 8.53 & 0.952 \\
    \bottomrule
  \end{tabular}
\end{table}

%% file: tables/methods_progression.tex
\begin{table}[htbp]
  \centering
  \caption{Methods progression for frozen DenseNet201 under the patient-level five-fold protocol (MAPE, \%): the global-average-pooling (GAP) baseline, the orientation-preserving grid-pooling head, Optuna tuning of that head, and the five-model out-of-fold ensemble. The single-model rows are per-fold CV means; the ensemble rows are computed on the pooled out-of-fold predictions (so no per-fold std is defined, ``--''). Aggregated the same pooled-OOF way, the tuned single DenseNet201 is \SI{10.14}{\percent} MAPE, so the genuine ensembling gain is \SIrange{10.14}{8.53}{\percent}; either way the ensemble breaks \SI{10}{\percent} on the cross-patient regime.}
  \label{tab:methods_progression}
  \begin{tabular}{lrrl}
    \toprule
    Method & MAPE (\%) & $\pm$std & Protocol \\
    \midrule
    DenseNet201, GAP head & 18.70 & $\pm$3.24 & patient 5-fold CV \\
    DenseNet201, grid pooling (frozen) & 12.59 & $\pm$2.15 & patient 5-fold CV \\
    DenseNet201, grid + Optuna-tuned & 10.80 & $\pm$2.59 & patient 5-fold CV \\
    \midrule
    Tuned 5-model ensemble (OOF mean) & 9.89 & -- & patient 5-fold OOF \\
    \textbf{Tuned 5-model ensemble (stacked)} & \textbf{8.53} & -- & patient 5-fold OOF \\
    \bottomrule
  \end{tabular}
\end{table}

%% file: sections/evaluation.tex
\section{Clinical-Grade Evaluation}
\label{sec:evaluation}

A point estimate of the Doppler angle is, on its own, of limited clinical use. The velocity correction scales as $1/\cos\theta$, so its sensitivity to angular error, $\partial(1/\cos\theta)/\partial\theta = \tan\theta/\cos\theta$, grows steeply toward the high angles used in carotid insonation. A sonographer therefore needs more than a predicted $\theta$: a calibrated sense of how far to trust it, and a check on whether the predictor sits systematically off the reading it is meant to assist. We address this with five analyses: distribution-free prediction intervals, agreement with the reference reading, rotation test-time augmentation, a classical orientation prior with learned--classical fusion, and gradient attribution. To describe the harder of the two sampling protocols of Section~\ref{sec:results}, all five are built on the \emph{patient-level} out-of-fold (OOF) predictions of the Optuna-tuned DenseNet201, with every volunteer-proxy group held out exactly once. Each number is recomputed from the saved predictions in \texttt{results/predictions/} by the \texttt{uda} evaluation modules. Table~\ref{tab:clinical} collects the headline values.

\input{tables/clinical}

\subsection{Distribution-free prediction intervals via conformal calibration}
\label{sec:conformal}

To attach a calibrated uncertainty to each prediction we use split-conformal prediction \citep{Vovk2005,Lei2018}, which turns any point predictor into interval estimates with no distributional assumptions. Residuals are scored on the signed, $180\degr$-wrapped scale appropriate to an orientation target. We take the $(1-\alpha)$ empirical quantile $q$ of the absolute residuals on a calibration set and emit the interval $\hat{\theta}\pm q$. The calibration and test halves are drawn to be patient-disjoint, a group-holdout split of $900$ calibration rows and $1{,}200$ held-out test rows in which no volunteer-proxy group spans both. Standard split-conformal coverage assumes exchangeable calibration and test points; a group-disjoint split breaks point-level exchangeability across the group boundary, so the relevant guarantee is the weaker group-conformal one, which holds at the level of groups and degrades when groups are few. With roughly six calibration groups here, the empirical quantile $q$ carries considerable sampling variance, and the coverage we report is one realization of a single seed-42 split rather than an average over folds.

\begin{figure}[htbp]
  \centering
  \includegraphics[width=0.62\linewidth]{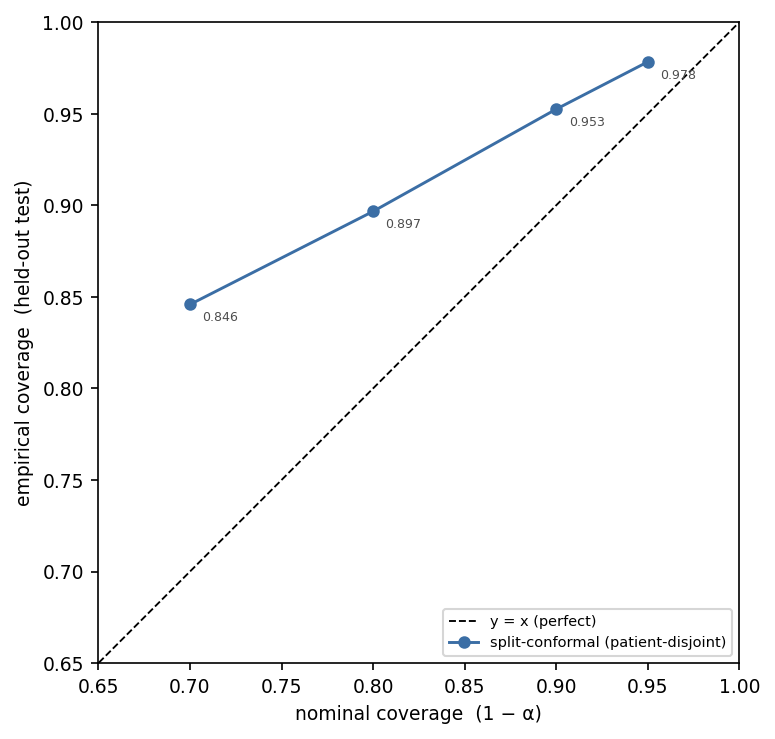}
  \caption{Conformal calibration on a single patient-disjoint calibration/test split:
  empirical coverage versus the nominal level, against the $y=x$ diagonal. The
  curve lies on or above the diagonal at every miscoverage level, consistent with
  valid coverage on this split; with only $\sim$six calibration groups, the
  empirical quantile carries large sampling variance and the result is not
  averaged over repeated splits.}
  \label{fig:calibration}
\end{figure}

On that split, empirical coverage meets or exceeds the nominal level at every miscoverage $\alpha$ we sweep. A nominal $90\%$ interval has half-width $\pm\SI{20.50}{\degree}$ and attains $95.2\%$ empirical coverage; a nominal $95\%$ interval ($\pm\SI{26.03}{\degree}$) reaches $97.8\%$. The over-coverage is the expected conservative behavior of group-conformal calibration on a small cohort (\num{12} patient-proxy groups), and because residuals live on the wrapped scale the half-widths never approach the $90\degr$ ceiling. Figure~\ref{fig:calibration} plots empirical against nominal coverage relative to the $y=x$ diagonal; the curve lies on or above the diagonal at the four levels swept. We read this as consistent with valid coverage on this split, not as a finite-sample proof: a single cal/test partition over six groups is one high-variance draw, and a different split could under-cover. We did not repeat the split, so the conformal result is held to a weaker standard than the patient-level cross-validation elsewhere in the paper, which reports $\pm1$~SD across folds.

One further caveat governs how the width should be read. The conformal residuals are scored over the rotated-view OOF set, whose effective sample size is closer to the handful of base images per group than to the $1{,}200$ row count, since the views are geometrically coupled copies (Section~\ref{sec:methods}). The interval is also built on the rotated-view residual distribution, whereas the headline accuracy reported below (Sections~\ref{sec:tta} and \ref{sec:fusion}) is a de-rotated, base-image quantity; the two are not on the same evaluation footing and should not be juxtaposed as a single number. With those limits stated, a $\pm\SI{20}{\degree}$ band under patient-level sampling is more informative than a bare point estimate, though at the steep insonation angles of Section~\ref{sec:intro} a band of this width propagates through $1/\cos\theta$ into a velocity-correction uncertainty of tens of percent, which we do not claim to have resolved.

\subsection{Agreement with the reference reading (Bland--Altman)}
\label{sec:bland_altman}

\begin{figure}[htbp]
  \centering
  \includegraphics[width=0.7\linewidth]{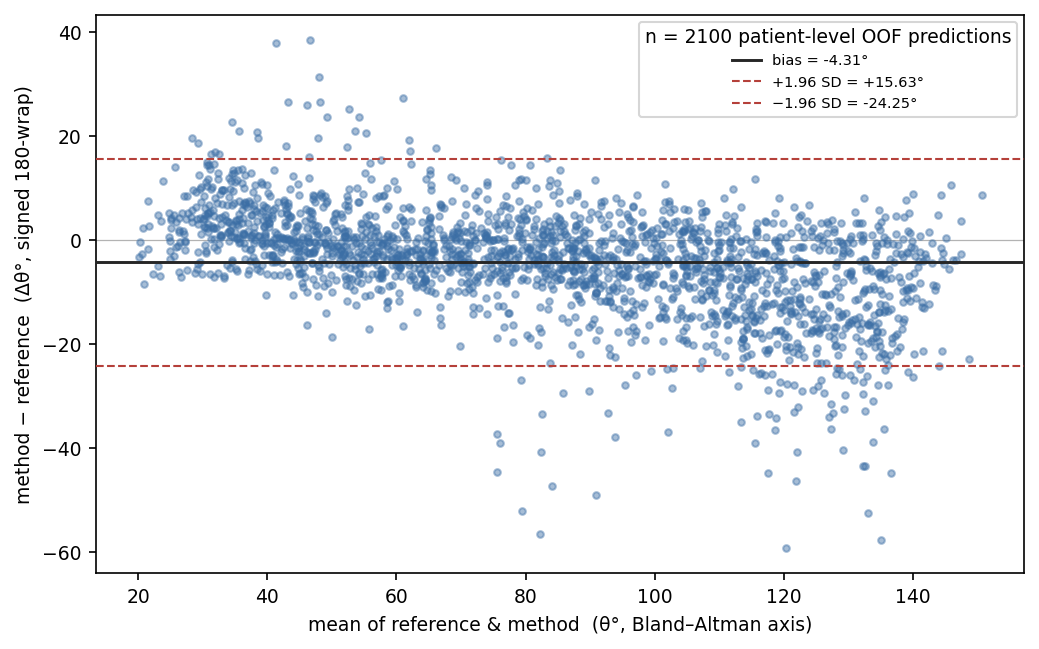}
  \caption{Bland--Altman agreement between the tuned model and the single
  reference reading, with mean bias and $95\%$ limits of agreement. This is a
  method-vs-reference comparison (one reading per image), not an inter-observer
  study. The $95\%$ limits are pooled over geometrically coupled rotated views and
  are therefore optimistically narrow.}
  \label{fig:bland_altman}
\end{figure}

To characterize systematic disagreement separately from aggregate error, we use a Bland--Altman analysis \citep{BlandAltman1986}, plotting the signed (wrapped) difference \emph{method minus reference} ($\theta_{\text{pred}}-\theta_{\text{true}}$) against their mean. Over the $2{,}100$ OOF predictions the model carries a small negative bias of $-\SI{4.31}{\degree}$ (it reads slightly lower than the reference) with $95\%$ limits of agreement of $-24.25$ to $+\SI{15.63}{\degree}$. These pooled limits treat the \num{25} coupled rotations of each base image as independent, so they are optimistically narrow; a repeated-measures correction would widen them. Aggregating each of the $n=12$ patient-proxy groups to a single value by a double-angle circular mean gives a comparable $-\SI{4.56}{\degree}$ bias with tighter limits, as the per-image scatter averages down. Figure~\ref{fig:bland_altman} shows no visible trend of the difference with the mean angle. We did not fit the regression of difference on mean (the standard proportional-bias test), so the constant-offset reading is a visual one rather than a quantified slope.

What this reference is matters for how the bias is interpreted. There is exactly one human reading per base image, the angle drawn in the MATLAB GUI, but the OOF residuals are scored against the applied-rotation label that defines each augmented view (Section~\ref{sec:methods}). The comparison is therefore method-vs-reference against a geometric construct, not an inter-observer study against a sonographer's clinical Doppler angle: there is one method (the model), one reference, no second annotator, and no repeated readings, so model error and the unmeasured variability of the human pipeline cannot be separated. The $-\SI{4.3}{\degree}$ bias should be read as the model's offset relative to that single annotation pipeline. Whether a constant $-\SI{4.3}{\degree}$ offset is benign for downstream velocity depends on the operating angle through the same $1/\cos\theta$ sensitivity, which we do not evaluate here. Acquiring repeated, multi-reader annotations to promote this into a genuine inter-observer analysis is the natural next step, and we record it as a limitation.

\subsection{Rotation test-time augmentation}
\label{sec:tta}

The orientation target invites a symmetry-aware test-time augmentation (TTA). Each of the 84 base images is presented at its 25 rotated views, every prediction is de-rotated back into the base frame, and the de-rotated estimates are reduced circularly, respecting the $180\degr$ period so that the reduction is seam-safe. Without TTA the base-image MAE is $\SI{7.80}{\degree}$ (RMSE $\SI{11.05}{\degree}$). A circular-mean reduction lowers it to $\SI{5.31}{\degree}$, and a circular-median reduction, the most robust of the three, reaches $\SI{4.72}{\degree}$ (RMSE $\SI{6.27}{\degree}$), a $39.5\%$ reduction in MAE at no training cost. The median is insensitive to the occasional badly oriented view, a useful safeguard in deployment. TTA composes with the conformal intervals of Section~\ref{sec:conformal}, sharpening the point estimate the intervals are wrapped around, though we note again that the TTA accuracy is a base-image quantity while the conformal width is computed over rotated views.

\subsection{A classical orientation prior and learned--classical fusion}
\label{sec:fusion}

We also situate the learned predictor against a purely classical baseline. A structure-tensor estimate of the dominant local orientation, a hand-built prior with no learning, reaches a base-image MAE of $\SI{3.16}{\degree}$. This figure is measured on the narrow native angle band of the 84 base images, a smaller-support and easier problem than the $\pm\num{60}\,\si{\degree}$ augmented range over which the conformal and Bland--Altman residuals are scored, so the numbers are not directly comparable. It does confirm that a usable geometric signal lives in the vessel-wall texture. Fusing the learned and classical estimates with a circular ($2\theta$, orientation-aware) combination reaches $\SI{2.72}{\degree}$ MAE. The fusion uses inverse-error weights; these were computed on the same 84 base images on which the fused estimate is scored, so the $\SI{2.72}{\degree}$ figure is in-sample and not a held-out result, and we report it as such pending a nested out-of-fold estimate. A fused estimate that beats both members is consistent with partly complementary error sources, but in-sample weight fitting and regression toward the mean of two correlated estimators can produce the same effect, so we do not claim more than that the two estimators carry some independent signal.

\subsection{Where the network looks (Grad-CAM)}
\label{sec:gradcam}

\begin{figure}[htbp]
  \centering
  \includegraphics[width=\linewidth]{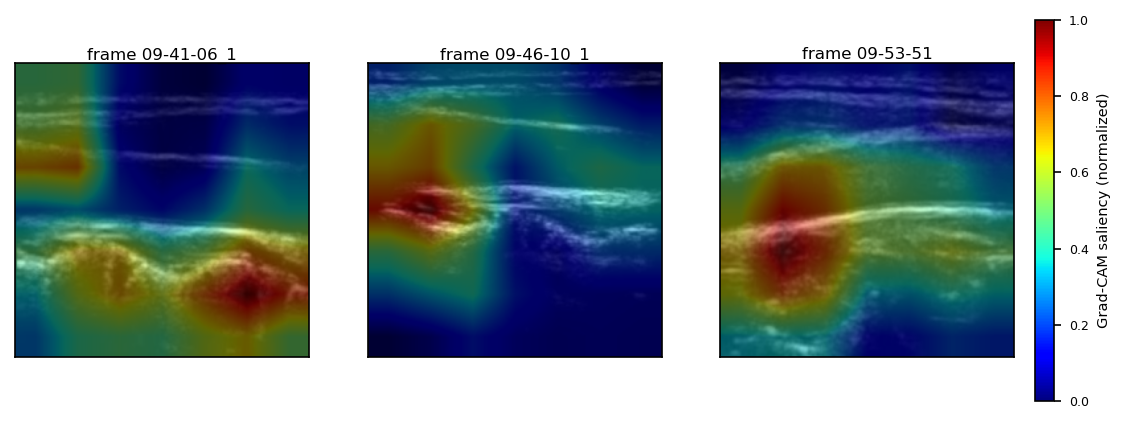}
  \caption{Grad-CAM attribution for the DenseNet201 backbone across three example
  frames (saliency normalized $0$--$1$; warm = high). In each frame the saliency
  concentrates on the vessel region rather than on speckle or image borders, and is
  diffuse rather than a pixel-thin trace of the wall, as expected for a coarse
  frozen feature grid. The evidence is illustrative from three frames, not a
  cohort-level attribution study.}
  \label{fig:gradcam}
\end{figure}

To check whether the predictor keys on anatomy rather than acquisition artifacts, we compute Grad-CAM saliency maps \citep{Selvaraju2017} for the DenseNet201 backbone. Figure~\ref{fig:gradcam} shows, across three example frames, saliency that tends to overlap the mid-image vessel band, the wall interfaces the structure-tensor prior of Section~\ref{sec:fusion} exploits, rather than the image borders or depth markers. The saliency is diffuse rather than a pixel-thin trace of the wall, as expected for the coarse $\sim\!8\times8$ feature grid of a frozen backbone, and in places it also lights up near-vessel tissue. This is anecdotal evidence from three frames on a coarse grid, so we read it cautiously: it shows where activation concentrates, which is on the vessel region, but it does not directly establish that the model encodes the orientation of the vessel boundary. The pattern is at least consistent with the model using the anatomical cue, and it is the reason the orientation-preserving grid-pooling head of Section~\ref{sec:methods} matters, since a rotation-invariant global pooling would discard the spatial layout this map makes visible.

\subsection{Summary}

These analyses give the patient-level estimator a partial clinical profile, with the caveats above attached to each piece. The conformal intervals show coverage at or above nominal on one patient-disjoint split, but on a single high-variance realization rather than a repeated-split estimate. The Bland--Altman bias is small ($-\SI{4.31}{\degree}$) and shows no obvious angle dependence by eye, against a single geometric reference rather than a second reader. The training-free TTA cuts base-image MAE by $39.5\%$, and an inverse-error fusion of the network with a classical prior reaches $\SI{2.72}{\degree}$ MAE in-sample. The attribution maps locate activation on the vessel region. The cohort is 84 images from $\sim\!10$ healthy young volunteers at a single center, with synthetic rotation labels and patient groups recovered from acquisition-time clustering rather than true subject IDs, so none of these results constitute external, multi-site, or pathology-validated evidence. End-to-end fine-tuning, transformer and medical foundation backbones, and self-supervised pretraining were beyond the Apple-silicon hardware used here; we report what we ran.

%% file: tables/clinical.tex
\begin{table}[htbp]
  \centering
  \caption{Clinical-grade evaluation of the tuned DenseNet201 on the patient-level OOF predictions. Split-conformal intervals achieve empirical coverage $\geq$ nominal; the Bland--Altman bias is small and \emph{negative} (model $-$ reference $<0$: the model reads slightly lower than the single reference reading); rotation TTA (median) nearly halves the raw base-image error. All numbers are recomputed from \texttt{results/predictions/} by the \texttt{uda} eval modules.}
  \label{tab:clinical}
  \begin{tabular}{llp{0.42\linewidth}}
    \toprule
    Quantity & Value & Detail \\
    \midrule
    \multicolumn{3}{l}{\emph{Split-conformal prediction intervals (patient-disjoint cal/test)}} \\
    90\% interval half-width & $\pm$20.50\degr & empirical coverage 95.2\% \\
    95\% interval half-width & $\pm$26.03\degr & empirical coverage 97.8\% \\
    \midrule
    \multicolumn{3}{l}{\emph{Bland--Altman, method $-$ reference (single reference reading)}} \\
    Bias & $-4.31\degr$ & 95\% LoA $-24.25$ \ldots\ $15.63\degr$ \\
    \midrule
    \multicolumn{3}{l}{\emph{Rotation test-time augmentation (base-image MAE)}} \\
    Raw (no TTA) & 7.80\degr & 84 base images \\
    Median TTA & \textbf{4.72\degr} & circular reduction over rotations \\
    \bottomrule
  \end{tabular}
\end{table}

%% file: sections/discussion.tex
\section{Discussion and Conclusion}
\label{sec:discussion}

\subsection{Orientation is the signal: why grid pooling is load-bearing}

The central deep-learning finding of this work is that reading the Doppler angle
$\theta$ from a B-mode image is an orientation-regression problem, and that the
conventional global-average-pooling (GAP) head is structurally unsuited to it.
GAP collapses a convolutional feature map by averaging over every spatial
position, a reduction that is approximately invariant to in-plane rotation of the
input. Invariance to orientation is the wrong inductive bias when the quantity to
be predicted is itself an orientation, because the head discards the very signal
the target depends on. The effect is large. Under image-level five-fold
cross-validation a frozen DenseNet201 with a GAP head scores
\SI{10.85}{\percent} MAPE, and under the stricter patient-level protocol it scores
\SI{18.7}{\percent}.

Replacing GAP with an orientation-preserving grid-pooling head recovers most of
that loss. The head average-pools the feature map onto a small $G\times G$ grid
and flattens it, so the coarse spatial layout of vessel structure survives into
the regressor. At image level this takes frozen DenseNet201~\citep{Huang2017} from
\SI{10.85}{\percent} to \SI{4.58}{\percent} MAPE under five-fold cross-validation,
roughly halving the error, and reaches \SI{5.84}{\percent} MAPE
(\SI{3.77}{\degree} MAE) on the original single augmented split
(Table~\ref{tab:replication}), reproducing the regime of the EMBC~2019
study~\citep{Patil2019} with no fine-tuning. The backbone, even frozen, already
encodes the vessel orientation; the work is in not pooling it away. The Grad-CAM
attributions~\citep{Selvaraju2017} (Fig.~\ref{fig:gradcam}) are consistent with
this reading, concentrating on the vessel region that defines the flow axis rather
than on background texture. We note that a hand-built structure-tensor prior reaches
\SI{3.16}{\degree} MAE on the narrow base-angle band with no learning at all
(\S\ref{sec:discussion}), so grid pooling is not the only route to a usable angle
estimate; what it provides is a frozen-backbone learned pipeline that is competitive
on the angular target. We offer the pooling lesson as a hypothesis worth testing
beyond this one task: when a regression target is a spatial pose or orientation, the
pooling operator should be chosen to preserve the geometry the label encodes rather
than average it out. The present evidence for that lesson is a single carotid task on
a few dozen subjects.

\subsection{Why newer backbones lose on tiny data}

The architecture bake-off (Fig.~\ref{fig:bakeoff}, frozen, patient-level five-fold
cross-validation, untuned heads) runs against the usual ImageNet-leaderboard
intuition. On the extraction-matched comparison, frozen DenseNet201 leads at
\SI{14.1}{\percent} MAPE ($\pm$\SI{2.19}{\percent}), ahead of the best modern
encoder, ConvNeXt-Base~\citep{Liu2022}, at \SI{15.7}{\percent}
($\pm$\SI{1.75}{\percent}); ConvNeXt-Tiny follows at \SI{16.1}{\percent} and the
EfficientNet and EfficientNetV2~\citep{Tan2019} families trail at
\SIrange{17}{21}{\percent}. The DenseNet and ConvNeXt-Base intervals overlap, so
the ranking among the leaders should be read as a draw rather than a clear win.
The remaining classic backbones (ResNet50~\citep{He2016},
Xception~\citep{Chollet2017}, InceptionV3~\citep{Szegedy2016},
VGG19~\citep{Simonyan2015}) were extracted with the replication-era pipeline, so we
do not rank them against ConvNeXt strictly like-for-like. The claim here is narrow:
on this task no newer or larger backbone improves on a frozen 2017 DenseNet201, not
that classic architectures beat modern ones as a family.

The likely reason is leverage rather than encoder quality. With only \num{84} base
images and frozen weights, no gradient reaches the backbone. The encoder supplies a
fixed feature basis, and the only fitted parameters live in the shallow head.
Architectures whose ImageNet advantage is realized through end-to-end training have
nothing to grip under a frozen protocol, and their feature statistics, tuned to a
classification objective, are no better matched to an angular regression target than
those of an older network. DenseNet201 happens to fit this task well across our
measurements: it yields the highest $R^2$, wins the replication, and remains
strongest after tuning. We read this as an empirical fact about our extraction
rather than a proven property of dense feature reuse, and we note that the original
study found VGG19 best under per-model tuning. We carry DenseNet201 forward. The
caution for small-cohort imaging is that benchmark rankings established on
million-image corpora do not transfer to a frozen-feature regime over a few dozen
subjects.

\subsection{Two protocols, two questions}

We report and tune the estimator under two complementary sampling protocols, which
answer different questions. \emph{Image-level sampling} partitions the
rotation-augmented corpus at the image level, the original study's protocol, and
measures how accurately the model reads $\theta$ across the population of
orientations and imaging conditions the augmentation spans. \emph{Patient-level
sampling} holds out whole volunteers via grouped cross-validation and measures
generalization to anatomy the model has never seen. The patient-level number is
naturally larger (Table~\ref{tab:dual_protocol}, Fig.~\ref{fig:protocol_gap}): once
volunteer-specific appearance can no longer be exploited, the model must transfer
across subjects. For the tuned stacked ensemble the image-level error is
\SI{2.79}{\percent} MAPE (\SI{1.96}{\degree} MAE, $R^2=0.995$) and the
patient-level error is \SI{8.53}{\percent} MAPE (\SI{5.93}{\degree} MAE,
$R^2=0.952$). The patient-level figure carries appreciable fold variance: per-fold
MAPE standard deviations are \SIrange{2.6}{4.4}{\percent} on the few held-out groups
per fold (Table~\ref{tab:dual_protocol}), so a spread of a few percentage points
around the \SI{8.53}{\percent} headline is expected, and the cross-subject estimate
should be read with that uncertainty in mind.

The gap between the two protocols is itself a measurement. It quantifies how much of
the within-population accuracy is anatomy-specific, that is, the cost of
cross-subject generalization. A reader sizing the method for deployment should weigh
the patient-level figure when the operative question is a new patient, and the
image-level figure when it is accuracy across the imaging conditions the corpus
represents. The per-backbone ordering is broadly stable across both lenses, though
not perfectly: under patient-level MAPE, VGG19 (\SI{14.97}{\percent}) and
InceptionV3 (\SI{14.66}{\percent}) sit within \SI{0.3}{} percentage points and
exchange ranks. The lesson for small-cohort medical imaging is that the sampling
protocol is part of the result. Augmentation manufactures correlated samples, so the
unit of statistical independence is the patient, not the image, and reporting both
protocols is more informative than privileging one.

Two comparisons need a word on aggregation, because the numbers are aggregated
differently. The single tuned DenseNet201 (\SI{3.00}{\degree} MAE,
\SI{4.03}{\percent} MAPE image-level) reproduces the best single model of the
EMBC~2019 work~\citep{Patil2019} ($\approx\SI{2.87}{\degree}$ MAE,
\SI{4.03}{\percent} MAPE). Head tuning~\citep{Akiba2019,Bergstra2011} and a
five-model stacked ensemble~\citep{Wolpert1992} improve on it and additionally
supply the previously unmeasured patient-level result. The per-backbone member
figures above are per-fold cross-validation means, whereas the ensemble figures are
pooled out-of-fold. Aggregated the same pooled-OOF way, the single tuned
DenseNet201 reaches \SI{7.80}{\degree} MAE patient-level, so the genuine ensembling
gain over the best single model is about \SI{1.9}{\degree} MAE; the larger apparent
gap between member and ensemble rows partly reflects the per-fold-mean versus
pooled-OOF aggregation rather than the ensemble alone.

\subsection{Uncertainty quantification and method-vs-reference agreement}

A point estimate alone is not clinically actionable, so we also report calibrated
prediction intervals. Split-conformal intervals~\citep{Vovk2005,Lei2018}, computed
on a single patient-disjoint calibration/test split, attain empirical coverage at or
above their nominal level at every level tested (Table~\ref{tab:clinical},
Fig.~\ref{fig:calibration}): the \SI{90}{\percent} interval has half-width
$\pm\SI{20.50}{\degree}$ with \SI{95.2}{\percent} empirical coverage, and the
\SI{95}{\percent} interval has half-width $\pm\SI{26.03}{\degree}$ with
\SI{97.8}{\percent} coverage. Two qualifications apply. The validity guarantee is
distribution-free and finite-sample under exchangeability; with grouped data and
only ten volunteers, exchangeability across patients is the assumption most at risk,
and coverage is marginal over the population rather than conditional on the
individual case. The reported coverages are single empirical numbers on a small test
set and therefore carry wide binomial uncertainty, so the observed over-coverage is
as consistent with conservative (somewhat too wide) intervals as with tight
calibration.

The intervals are wide, and the width is a property of a ten-volunteer cohort rather
than a modelling artifact. A wide interval that covers is still more useful for a
velocity-correction decision than a narrow one that does not, because it does not
mask the state of the evidence. We are careful about what such an interval offers in
practice: because the half-width is fixed across cases, it reports population-level
coverage and does not flag which individual readings are unreliable. For the
$\cos\theta$ velocity correction a half-width of \SIrange{20.50}{26.03}{\degree} can
move the correction factor markedly, so whether a band this wide is directly
actionable per case is an open question that external data would have to settle.

The Bland--Altman analysis~\citep{BlandAltman1986} (Fig.~\ref{fig:bland_altman})
reports a bias of $-\SI{4.31}{\degree}$ (method minus reference: the model reads
slightly low) and limits of agreement of $-24.25$ to $+\SI{15.63}{\degree}$. There
is exactly one human angle reading per image, the MATLAB-GUI annotation that serves
as the label, so this is a method-vs-reference comparison against a single reading
and not an inter-observer study. Without a second reader we cannot quantify the
irreducible human labelling variability against which an automated method should
ultimately be judged, and the reference reading is itself a noisy estimate of the
true flow angle. The bias also has a direction. A consistent low-angle bias shifts
the $\cos\theta$ velocity correction the same way for every case, which for an
application like stenosis grading is a systematic error rather than random scatter,
and a directional offset of this size warrants attention even though it is small and
fully characterized in our cohort.

Two further analyses are encouraging within these bounds, though on a different
evaluation unit. Both are computed on the \num{84} base images over the narrow
base-angle band, not on the patient-level OOF protocol, so they are not directly
comparable to the patient-level ensemble MAE of \SI{5.93}{\degree}. Rotation-based
test-time augmentation, combined under a seam-safe circular median over de-rotated
views, reduces the raw base-image MAE from \SI{7.80}{\degree} to \SI{4.72}{\degree}
with no retraining. A circular fusion of the learned estimate with a classical
structure-tensor angle prior reaches \SI{2.72}{\degree} MAE on the same band,
beating either the learned model or the geometric prior (\SI{3.16}{\degree} MAE)
alone, which suggests that a data-driven encoder and a hand-crafted geometric cue
capture partly complementary information.

\subsection{Clinical relevance and deployment}

The clinical appeal of an image-only estimator is that it slots into the existing
Doppler workflow without changing it. Angle correction is today a manual cursor
placement, a recurring source of error and, as noted in \S\ref{sec:intro}, a
deficiency flagged in vascular-laboratory accreditation. An automated reading
accurate to a few degrees could pre-populate or sanity-check that cursor and reduce
the operator-dependent variation in the measured angle. We are careful not to
overstate this. The labels come from a single annotator with no inter-observer
baseline, so an estimator trained against them can at best reproduce that
annotator's readings and their biases; it cannot independently correct operator
error or claim to standardize against a validated ground truth. Any downstream
benefit to velocity estimation or stenosis grading is a plausible motivation rather
than a measured result, since neither was evaluated here.

Because the estimator consumes only the formed grayscale B-mode frame, with no
color Doppler, no segmentation, and no knowledge of the vendor-specific beamforming
path, it could in principle be packaged as a software add-on. We have not tested
this: the cohort is single-scanner and single-center, so cross-vendor behavior is
unknown and the device-agnostic framing is a design property rather than a validated
capability. The conformal intervals are relevant to deployment in that a tool
reporting a calibrated band alongside its point estimate communicates population-level
uncertainty rather than a bare number, which is a safer default than silent
over-confidence. As noted above, a fixed-width marginal interval does not identify
which individual cases are uncertain, so it supports a general posture of caution
more than per-case triage. Realizing any of this would require the external
validation discussed below; the present results establish that the core estimate is
accurate and reproducible under our protocols, which is what justifies pursuing that
validation.

\subsection{Limitations}

The cohort is small: about ten volunteers and \num{84} base images from a single
center, a single scanner, and the carotid only. Patient-level cross-validation
therefore rests on a handful of held-out subjects per fold, the per-fold variance is
non-negligible (see the \SIrange{2.6}{4.4}{\percent} MAPE standard deviations of
Table~\ref{tab:dual_protocol}), and external validity to other vendors, anatomies,
and acquisition settings is untested. The method has never seen a tortuous vessel, a
bifurcation, plaque, or a non-carotid site, so any workflow claim beyond the imaged
population is speculative. There is a single reference reading per image, so the
agreement analysis is method-vs-reference rather than inter-observer, and the upper
bound on achievable accuracy set by human labelling noise is unknown. Finally, every
learned result here uses a frozen ImageNet backbone with only a trained pooling head;
the encoder is never updated. The estimator should be read as a strong, fully
reproducible frozen-feature baseline, not as the ceiling of what this task admits.

\subsection{Future work}

The frozen-only scope leaves several directions for once dedicated accelerator
hardware is available. End-to-end fine-tuning of the backbone, left unrun here, is
the most direct candidate for improving the patient-level number, since it would let
the encoder adapt its features from a generic classification basis to the
angular-regression objective. Self-supervised pretraining on the unlabelled carotid
frames could yield representations less dependent on the small labelled cohort and so
better cross-subject transfer. Geometry-aware alternatives to grid pooling, an
explicitly circular treatment of the angular target across the full $360^{\circ}$
range, and orientation-equivariant heads are natural architectural extensions of the
pooling insight. On the evaluation side, a multi-reader study would convert the
method-vs-reference agreement into a proper inter-observer comparison and establish
the human-noise floor, and a multi-center, multi-vendor cohort would test the
external validity that a single-center dataset cannot. None of these is claimed as a
result of the present paper; they are the experiments the present results motivate.

\subsection{Conclusion}

We have presented a reproducible, frozen-backbone estimator of the Doppler
beam-to-vessel angle from a single grayscale B-mode carotid image, and isolated the
design choice that makes a frozen pipeline competitive on this orientation-regression
task: orientation-preserving grid pooling in place of global average pooling. Optuna
head tuning and a five-model stacked ensemble yield our best estimator, reported
under two complementary sampling protocols. The image-level error is
\SI{2.79}{\percent} MAPE (\SI{1.96}{\degree} MAE, $R^2=0.995$) and the patient-level
error is \SI{8.53}{\percent} MAPE (\SI{5.93}{\degree} MAE, $R^2=0.952$), the latter
with the fold variance noted above. Against the one prior point of comparison on this
task, the EMBC~2019 study, the tuned ensemble improves on the published image-level
figures. An architecture bake-off shows that no newer or larger backbone improves on
a frozen 2017 DenseNet201 on this cohort, and the evaluation equips the method with
distribution-free conformal intervals and a method-vs-reference agreement analysis.
The takeaway is twofold. For this task, automated Doppler-angle estimation is
accurate and now reproducible, characterized under both the image-level and the
stricter patient-level lens. For small-cohort medical imaging more broadly, matching
the pooling operator to the geometry of the target, reporting accuracy under more
than one sampling protocol, and supplying calibrated intervals rather than a single
point estimate are practices that make such a method easier to trust.

\subsection*{Reproducibility}

Every table and figure in this paper is regenerated from saved results by code, with
no hand-entered numbers. The data is the public SPLab common-carotid B-mode set; the
pipeline (data preparation, the rotation-augmentation corpus, frozen feature
extraction, the grid-pooling head, Optuna tuning, ensembling, and the conformal /
Bland--Altman / TTA evaluation) is deterministic under a fixed seed ($42$) and runs
on a CPU. After preparing the corpus and cross-validation results, the analysis
tables are produced by a single table-generation script and the figures by a single
figure-generation script; both read only the saved cross-validation summaries and
out-of-fold predictions. The frozen patient- and image-level five-fold
cross-validation that underlies the protocol comparison and the architecture bake-off
is likewise produced by a single command per backbone, so the reported means and
standard deviations can be reproduced end to end from the raw images.